%% file: main.tex
\documentclass[sigconf]{acmart} 

\usepackage{amsmath}

\usepackage{multirow}
\usepackage{adjustbox}

\usepackage{pifont}
\usepackage{bbding}

\usepackage{algorithm}
\usepackage{algpseudocode}
\usepackage{setspace}

\usepackage{xcolor}

\usepackage[normalem]{ulem}
\useunder{\uline}{\ul}{}

\usepackage[utf8]{inputenc} 
\usepackage[T1]{fontenc}    
\usepackage{hyperref}       
\usepackage{url}            
\usepackage{booktabs}       
\usepackage{amsfonts}       
\usepackage{nicefrac}       
\usepackage{microtype}      
\usepackage{xcolor}         

\usepackage{graphicx}
\usepackage{subfigure}

\usepackage{enumitem}

\AtBeginDocument{%
  }

\begin{document}

\author{Tingyue Pan$^{1}$, Mingyue Cheng$^{1\ast}$, Shilong Zhang$^{1}$, \\  Zhiding Liu$^{1}$, Xiaoyu Tao$^{1}$, Yucong Luo$^{1}$, Jintao Zhang$^{1}$, Qi Liu$^{1}$}

\affiliation{%
	\institution{$^1$State Key Laboratory of Cognitive Intelligence, University of Science and Technology of China, Hefei, China}
	\country{}}




\email{{pty12345,zhangshilong,zhiding,txytiny,prime666,zjttt}@mail.ustc.edu.cn, {mycheng,qiliuql}@ustc.edu.cn}





\title{OneCast: Structured Decomposition and Modular Generation for Cross-Domain Time Series Forecasting}

\renewcommand{\shortauthors}{Tingyue Pan et al.}

\begin{abstract}
Cross-domain time series forecasting is a valuable task in various web applications. Despite its rapid advancement, achieving effective generalization across heterogeneous time series data remains a significant challenge. Existing methods have made progress by extending single-domain models, yet often fall short when facing domain-specific trend shifts and inconsistent periodic patterns. We argue that a key limitation lies in treating temporal series as undifferentiated sequence, without explicitly decoupling their inherent structural components. To address this, we propose OneCast, a structured and modular forecasting framework that decomposes time series into seasonal and trend components, each modeled through tailored generative pathways. Specifically, the seasonal component is captured by a lightweight projection module that reconstructs periodic patterns via interpretable basis functions. In parallel, the trend component is encoded into discrete tokens at segment level via a semantic-aware tokenizer, and subsequently inferred through a masked discrete diffusion mechanism. The outputs from both branches are combined to produce a final forecast that captures seasonal patterns while tracking domain-specific trends. Extensive experiments across eight domains demonstrate that OneCast mostly outperforms state-of-the-art baselines. Our code is publicly available \footnote{\url{https://github.com/pty12345/OneCast}}.



\end{abstract}








\settopmatter{printacmref=false}

\maketitle

\section{Introduction}

Time series forecasting (TSF) serves as a fundamental technique in many web applications such as traffic scheduling \cite{casado2021web}, e-commerce systems \cite{hu2015web}, and website maintaining \cite{syu2021qos}. While recent advances have boosted forecasting accuracy \cite{cheng2024convtimenet, wang2024timexer,  cheng2025comprehensive}, most efforts focus on single domains. In practice, real-world web signals from different platforms exhibit transferable temporal regularities, motivating cross-domain forecasting to leverage such shared temporal dynamics for more accurate prediction \cite{liu2024unitime, cheng2025cross, ekambaram2024tiny}.



\vspace{13mm}

\begin{figure}[htbp]
    \centering
    \includegraphics[width=1.0\linewidth]{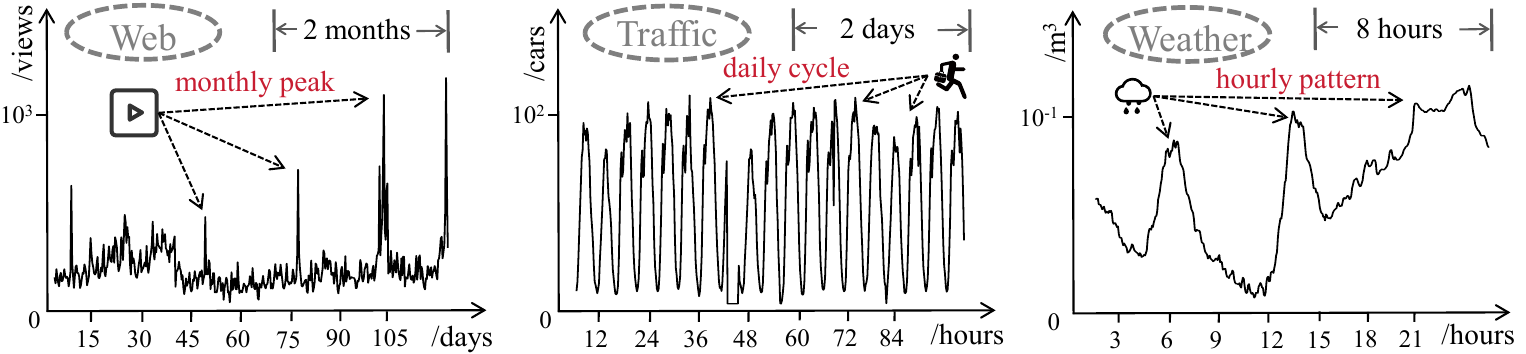}
    \caption{Time series data from different domains exhibit significant heterogeneity, such as variations in different, numeric ranges, sampling rates and key patterns.}
    \label{fig:genious}
\end{figure}

Despite their potential, cross-domain time series forecasting remains challenging due to the inherent data heterogeneity (shown in Figure \ref{fig:genious}). At the data-format level, different temporal series vary in sampling frequency, numerical scales, and measurement units, with additional discrepancies in data resolution, missing-value distribution, and normalization conventions. These inconsistencies make even value alignment across domains difficult. At the sequence-structure level, temporal series exhibit specific trend and seasonality characteristics across domains. First, trend dynamics are highly domain-specific, ranging from steady growth or decline to irregular nonlinear fluctuations, which hinder the direct transferability of temporal representations. Second, seasonal patterns differ in frequency and phase (e.g. yearly climate cycles v.s. daily traffic rhythms), which undermines a single shared notion of periodicity. These challenges call for approaches that explicitly address data format and structural discrepancies while enabling effective knowledge sharing across domains.

To address the above challenges, a variety of studies have been proposed. Alignment-based methods learn unified representations in continuous space to bridge sampling inconsistencies, yet they often simplify multi-variable sequence into a single channel without modeling inter-series dependencies \cite{das2024decoder, ekambaram2024tiny}. Discretization-oriented methods instead convert raw values into codebook tokens or normalized buckets, thereby reducing peak value and local-pattern discrepancies across domains \cite{talukder2024totem, ansari2024chronos}. Despite these advances, existing approaches still treat time series as undifferentiated sequence, lacking explicit decomposition into trend and seasonal structures. As a result, they struggle to capture domain-specific trend shifts and misaligned periodic patterns, which ultimately constrains their ability to generalize across heterogeneous domains.


In our view, an effective cross-domain forecasting model not only reconcile data formats but also perform structural adaptation, which is important in improving cross-domain generalization and interpretability. Building on this insight, we propose OneCast, which couples structured decomposition with modular generation to realize component-specific forecasting. Specifically, OneCast decompose time series into seasonal and trend parts. For seasonality with transferable periodic patterns across domains, OneCast employs a lightweight projection module that maps inputs onto interpretable basis functions (e.g. daily or weekly harmonics). Consequently, this design yields accurate seasonality reconstruction with clear interpretability . For the trend component that exhibits disparity across domains, OneCast encodes segment-level patterns into discrete tokens via a semantic-aware tokenizer trained with innovated dual-decoder strategy. This yields pattern-level representations that are less sensitive to peak value and distribution shift. To model the evolution of these tokens through a stochastic refinement process, we train a diffusion-based token predictor by masked strategy, and then iteratively generate future tokens when inference, capturing long-range dependencies and enabling decoding of high-confidence tokens. Finally, OneCast fuses the frequency-reconstructed seasonality and the diffusion-generated trend into the integrated final forecast, thereby facilitating effective knowledge coordination across heterogeneous sources within a unified formulation. Our contributions are as follows:

\begin{itemize}  [leftmargin=*]
\item We propose structured decomposition of seasonality and trend coupled with modular generation to mitigate periodicity mismatches and trend shifts, improving cross-domain generalization and interpretability.

\item We introduce a dual-decoder training strategy that mitigates the history–future distribution shift, coupling reconstruction and forecasting supervision for the trend tokenizer. 

\item We develop a diffusion-based token predictor that iteratively generates tokens in confidence-awareness, which captures long-range dependencies and alleviates error accumulation.

\end{itemize}

\section{Related Work}

\textbf{Time Series Forecasting.} Time series forecasting plays a vital role in many real-world applications \cite{casado2021web,cheng2025comprehensive}. Traditional statistical methods such as ARIMA \cite{hyndman2008forecasting} rely on linear and stationary assumptions, offering interpretability but limited capacity for nonlinear dynamics. With the advent of machine learning, models like XGBoost \cite{zhang2021time} and LightGBM \cite{ke2017lightgbm} enhance nonlinear modeling but still lack temporal representation learning. Deep learning approaches further advanced this field by capturing complex temporal and cross-channel dependencies in multivariate time series. Among these, temporal decomposition modeling, which boasts the advantage of decomposing complex time series into interpretable sub-components, has become a common part in time series modeling. For example, Transformer-based architectures like Autoformer \cite{wu2021autoformer} and Fedformer \cite{zhou2022fedformer} using deep decomposition architectures to iteratively extract more predictable components for future data modeling. Beyond them, architectures like CNN-based TimesNet \cite{wu2022timesnet} and MICN \cite{wang2023micn}, MLP-based DLinear \cite{zeng2023transformers} also demonstrate promising forecasting performance. More recently, reasoning-capable language models further show their potential in extending forecasting tasks through in-context learning and step-by-step reasoning \cite{luo2025time,wang2025can}.

\textbf{Cross-domain Modeling. } 
In recent years, extensive research on cross-domain modeling has been widely explored \cite{liu2024unitime, cheng2025cross}. In the representation-alignment paradigm, UniTime \cite{liu2024unitime} segments sequence into patches and adopts patch embedding spaces as unified input representation spaces. Going a step further, TimesFM \cite{das2024decoder} maintains frequency-specific embedding dictionaries to enhance the generalization ability across different sampling rates. In the discretization-oriented paradigm, methods such as Chronos \cite{ansari2024chronos} and TOTEM \cite{talukder2024totem} leverage quantiles of the overall distribution to map each time point or slice to discrete IDs. On this basis, PromptCast \cite{xue2023promptcast} represents numerical sequence as text and exploits reasoning ability of large language models for forecasting. At foundation level, Moirai \cite{liu2024moirai} predicts distributions and aligns domains via divergence, but requires careful selection of the prior probability function. Despite these advances, most methods still operate directly on raw sequence without explicitly disentangling trend and seasonal components, which hampers principled transfer of structural temporal knowledge across domains.

\section{Methodology}

\begin{figure*}[htbp]
    \centering
    \includegraphics[width=\linewidth]{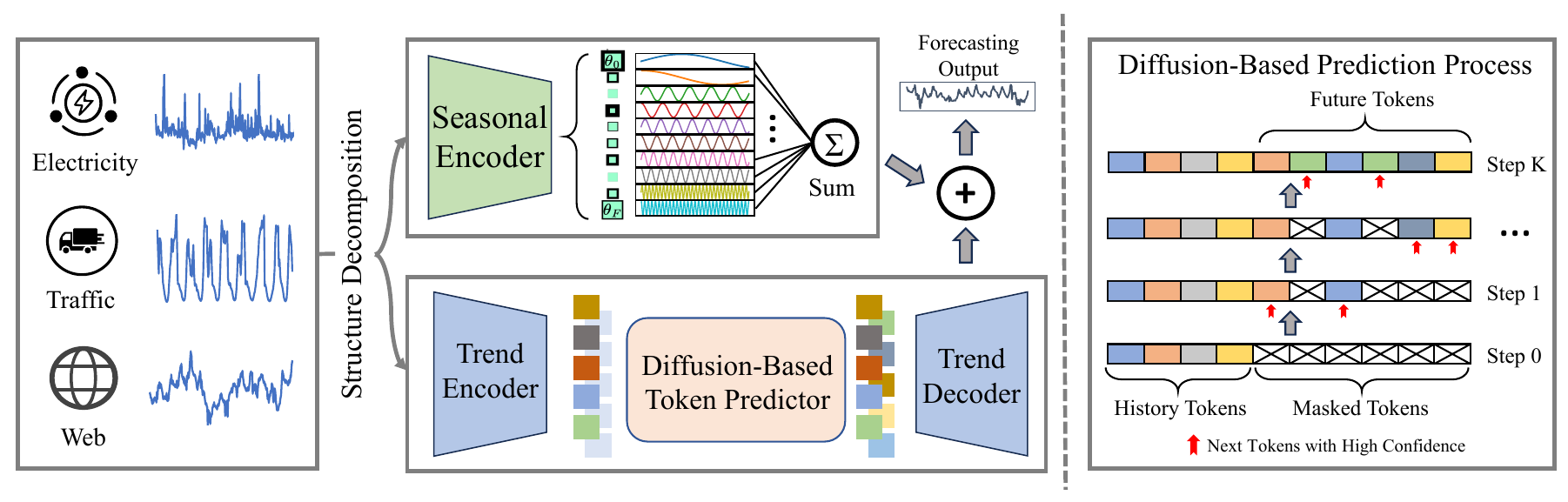}
    \caption{The inference pipeline of OneCast. Right: detailed prediction process of diffusion-based token predictor.}
    \label{fig:overall_framework}
\end{figure*}

In this section, we will present the detailed design of OneCast. As illustrated in Figure \ref{fig:overall_framework}, the main architecture of OneCast consists of two components: 1) a seasonal prediction module designed to estimate the weights of predefined periodic functions, and
2) a trend prediction module that equipped with unified discrete tokenizer and diffusion-based confidence-aware token predictor. 

\subsection{Problem Definition}
We focus on the cross-domain time series forecasting task. For each time step \( t \), the multivariate time series observed in domain \( d \) is defined as \( \mathbf{x}_t^d = \{x_t^{d,1}, \ldots, x_t^{d,c_d}\} \in \mathbb{R}^{c_d} \), where \( c_d \) denotes the channels of domain \( d \).
Assuming the length of the historical window and future window is \( L_h \) and \( L_f \) respectively, the historical series is \( \mathbf{X}_{L_h}^d = \{\mathbf{x}^d_1, \ldots, \mathbf{x}^d_{L_h}\} \) and the forecasting target (i.e., the future series) is \( \mathbf{X}_{L_f}^d = \{\mathbf{x}^d_{L_h+1}, \ldots, \mathbf{x}^d_{L_h+L_f}\} \).
The objective is to learn a unify model that can predict the future time series \( \hat{\mathbf{X}}^d_{L_f} \) based on the historical observations \( \mathbf{X}^d_{L_h} \) across different domains.



\subsection{Seasonal Prediction Module} The seasonal component typically exhibits clear periodic information, which can be represented as a weighted sum of a series of periodic functions approximately \cite{oreshkin2019n, darlow2024dam}. Specifically, for the seasonal component \( X_S \) with length \( T \), the i-th feature can be approximately expressed as:

\begin{equation}
    X_S^i(t) = \sum_{j=1}^{N_s} [ v_{j,s}^i \cdot \sin(w_j t) + v_{j,c}^i \cdot \cos(w_j t) ] ,
    \label{eq:season}
\end{equation}

\noindent where \( t \in \{1, 2, \ldots, T\} \) denotes the time step, \( N_s \) is the number of selected periodic function, \( w_i \) represents the \( i \)-th frequency component, and \( v_{j,s}^i \) denotes the weight corresponding to the frequency component. We regard each periodic function as a code word, with \( w_i \) initialized based on common periodicity (e.g. one day, one week, etc.) \cite{darlow2024dam}. Since all available periodic components are fixed, it suffices to determine the weights corresponding to each periodic component for different seasonal components, without cumbersome selection of which periodic components to use. However, these weights are not static and should be dynamically and adaptively generated based on the sequence's context. Therefore, we utilize a simple multi-layer perceptron (MLP) to learn this mapping and predict the weights for future sequence.

\subsection{Trend Prediction Module}

In this section, we mainly introduce the modeling of the trend component. By summarizing the limitations of previous studies, we propose to decouple this modeling task into two modules: a tokenizer module and a prediction module.

\subsubsection{Semantic Tokenizer} We consider employing vector quantization (VQ) technique \cite{van2017neural} to learn the mapping from continuous trend data to semantic discrete tokens in a data-driven manner. Its core idea is to assign a unique identifier code to each local region of the subsequence through reconstruction optimization based on the autoencoder architecture. Such a specific code is selected from a predefined codebook embedding space \cite{van2017neural}. And to prevent the index collapse in training process of VQ, we introduce a learnable linear transformation matrix \( M \) to ensure all codes are activated during parameter updating \cite{zhu2024addressing}. 

Formally, we assign a trainable vocabulary \( E = \{ \mathbf{e}_1, \mathbf{e}_2, \ldots, \mathbf{e}_K \} \in \mathbb{R}^{K \times D} \) consisting of \( K \) distinct \( D \)-dimensional vectors. The indices of these vectors can be regarded as discrete tokens for the time series. Assuming the i-th domain input sequence to the encoder is \( X_i \), the encoded temporal feature is \( Z_i = \{ \mathbf{z}_1, \mathbf{z}_2, \ldots, \mathbf{z}_n \} \in \mathbb{R}^{n \times D} \), where $n$ denotes the number of temporal tokens. Before applying the nearest neighbor strategy for matching, a matrix \( M \in \mathbb{R}^{D \times D} \) is used to linearly transform the vocabulary space, i.e., \( \hat{E} = E \cdot M \), to activate all codes. Subsequently, for each \( \mathbf{z}_i \), the nearest neighbor strategy replaces \( \mathbf{z}_i \) with the transformed vector \( \hat{\mathbf{e}}_k \), where \( k = \arg\min_k \|\mathbf{z}_i - \hat{\mathbf{e}}_k\| \). The quantilized output is the resulting discrete token sequence \( \hat{S} = \{k_1, k_2 \ldots k_n\}\).

To address the non-differentiable gradient issue caused by the nearest neighbor replacement, the VQ network introduces a straight-through estimator, with the loss function shown below:

\begin{equation}
    \mathcal{L}_{codebook} = \|sg[\mathbf{z}_i] - \hat{\mathbf{e}}_k\|^2 + \beta \|\mathbf{z}_i - sg[\hat{\mathbf{e}}_k]\|^2 ,
    \label{eq:codebook}
\end{equation}
\vspace{-0.1in}

\noindent where $sg[\cdot]$ denotes the stop gradient operator, and \(\beta\) is the hyperparameter controlling the relative learning speed of the encoder.

\label{token_predictor}
\subsubsection{Diffusion-based Token Predictor}

We introduce a confidence-aware discrete diffusion generation paradigm \cite{ho2020denoising} to generate future trend token. As shown in the right of Figure \ref{fig:overall_framework}, the discrete diffusion-based token predictor generates multiple tokens at once and employs a full attention mechanism to greedily select the next batch of tokens based on generation probabilities. In this way, the model overcomes the constraints of fixed causal generation, and enables parallel decoding of high-confidence tokens while mitigating autoregressive error accumulation.

Specifically, given a historical trend token sequence $T_h$, we initialize the future sequence $\tilde{T_f}$ by replacing all positions with [mask], forming the input $\tilde{T_t}=Concat(T_h, \tilde{T_f})$. The generation proceeds through $K$ denoising steps. At each step, the model restores a fixed number of masked tokens by selecting the positions with the highest confidence scores and filling them with the most probable candidates. The updated sequence is then fed back as input for the next round, progressively reducing the number of masked tokens. After $K$ iterations, all masked positions are resolved, yielding the complete sequence of predicted future trend tokens $\hat{T_f}$.

\begin{figure*}[t]
    \centering
    \includegraphics[width=\textwidth]{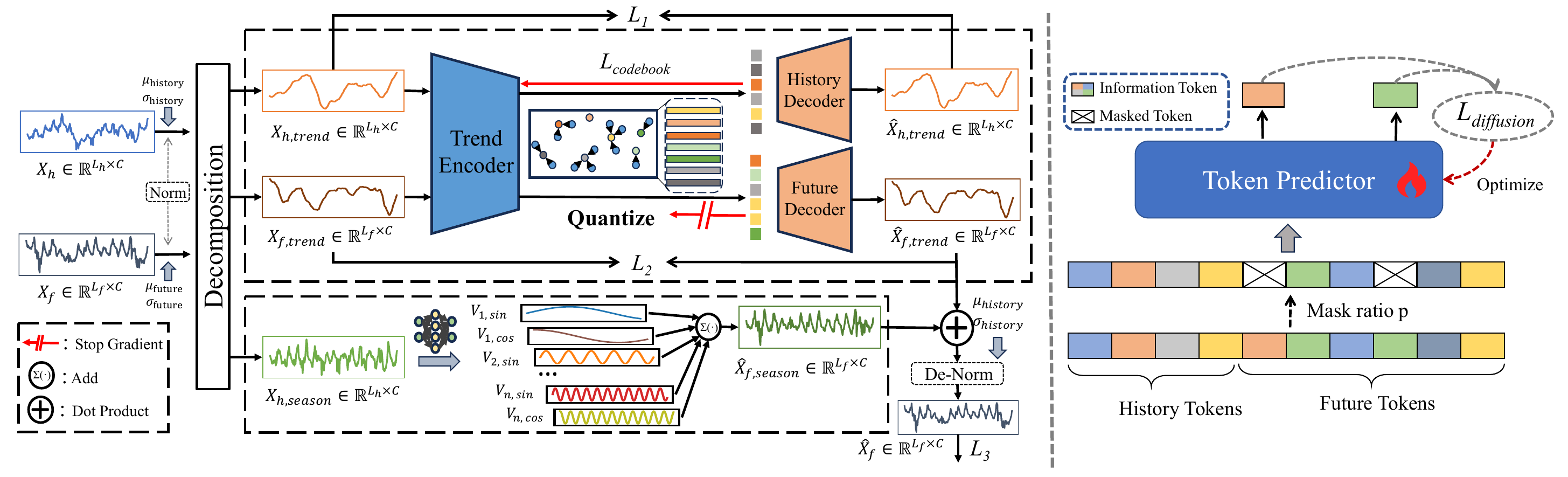}
    \caption{The complete training pipeline of OneCast. Left: joint optimization of seasonal predictor and semantic trend tokenizer, where $X_{h}$ and $X_f$ represents the history and future windows, respectively. Right: training process of confidence-aware diffusion-based token predictor. } 
    \label{fig:tokenizer}
\end{figure*}

\subsection{Optimization of OneCast}

Given the interactions between seasonal and trend components, we pursue end-to-end optimization. However, the trend token predictor depends on stable token representations, making it unsuitable for direct joint training with the tokenizer. To this end, we adopt a two-stage strategy: 1) joint optimization of the seasonal predictor and trend tokenizer end-to-end, and 2) independent training of diffusion-based token predictor on stabilized representation space.


\subsubsection{Training Stage \uppercase\expandafter{\romannumeral1}} Left of Figure \ref{fig:tokenizer} illustrates the joint optimization of both seasonal predictor and semantic trend tokenizer. For clarity, we denote the time series segment sampling from the \(i\)-th domain with $C$ channels as \(X \in \mathbb{R}^{L \times C}\), and divide it into a historical window \(X_h \in \mathbb{R}^{L_h \times C}\) and a future window \(X_f \in \mathbb{R}^{L_f \times C}\) without overlap. To mitigate the issue of distribution shift and narrow the sample distribution range that the tokenizer needs to encode, we apply instance normalization \cite{kim2021reversible} to \(X_h\) and \(X_f\):

\vspace{-0.5mm}

\begin{equation}
    X_{h,norm} = \frac{X_{h} - \mu_{h}}{\sqrt{\sigma^2_{h} + \epsilon}} ,     X_{f,norm} = \frac{X_{f} - \mu_{f}}{\sqrt{\sigma^2_{f} + \epsilon}},
\end{equation}
\vspace{-0.1in}

\noindent where $\epsilon$ is a small constant used to prevent division by zero errors, $\mu_{h}, \mu_{f}, \sigma_{h}$ and $ \sigma_{f} \in \mathbb{R}^{C}$ are the mean and variance of historical and future windows in different channels, respectively. Subsequently, we perform the moving average on \(X_{h,norm}\) and \(X_{f,norm}\), obtaining trend components \(X_{h,trend}\) and \(X_{f,trend}\), as well as seasonal components \(X_{h,season}\) and \(X_{f,season}\).


Despite the use of instance normalization, the lack of true mean and variance for the future window implies that only the statistics from historical window can be utilized for denormalization when reconstructing continuous series with future tokens. This may result in statistical discrepancies in the future window, which can hinder a single decoder-based VQ-VAE from accurately reconstructing future tokens to the true scale of the future window.

To enable the tokenizer to utilize the statistics of historical window in aiding the reconstruction of future tokens, we introduce a novel dual-decoder design when training. Specifically, the first decoder, $\mathbf{D}_h$, is responsible for sequence compression optimized by reconstructing historical trend $\hat{X}_{h,trend}$, while the second decoder $\mathbf{D}_f$, leverages the mean and variance of the historical window to aid in decoding future tokens to $\hat{X}_{f,trend}$. This design allows to effectively integrate instance normalization techniques, thereby decoding future tokens more precisely to the scale of future window. We present the optimization objective $ \mathcal{L}_{1} $ for trend component compression and $ \mathcal{L}_{2} $ for future tokens decoding as follows:

\begin{equation}
    \begin{aligned}
        \mathcal{L}_{1} &= \|\mathcal{D}_{(\mu_h,\sigma_h)}{X_{h,trend} - \mathcal{D}_{(\mu_h,\sigma_h)}\hat{X}_{h,trend}}\|^2 ,\\
        \mathcal{L}_{2} &= \|\mathcal{D}_{(\mu_f,\sigma_f)}X_{f,trend} - \mathcal{D}_{(\mu_h,\sigma_h)}\hat{X}_{f,trend}\|^2 ,
    \label{eq:reconst}
    \end{aligned}
\end{equation}



\noindent where $\mathcal{D_{\mu,\sigma}(\cdot)}$ refers to the process of denormalization using the corresponding mean and variance statistics for the sequence. $\mathcal{L}_{1}$ encourages the tokenizer to learn the ability to compress sequence, while $\mathcal{L}_{2}$  compels it to utilize statistics from historical windows for decoding future tokens. It is worth noting that the gradients introduced by $\mathcal{L}_{2}$ are truncated before being backpropagated to the codebook. This prevents them from interfering with the learning of effective token compression.

Incorporating the optimization of codebook from Equation \ref{eq:codebook}, the overall trend tokenizer update function $\mathcal{L}_{trend}$ is as follows:

\begin{equation}
    \mathcal{L}_{trend} = \mathcal{L}_{1} + \mathcal{L}_{2} + \mathcal{L}_{codebook}~.
\end{equation}

For the seasonal component, according to Equation \ref{eq:season}, we use historical seasonal series $X_{h,season}$ to predict the weights of future seasonal components $\{v_{j,c}^i\} \in \mathbb{R}^{N \times C}$, through a multilayer perceptron, thus calculating the future seasonal terms $\hat{X}_{f,season}$. Furthermore, we are able to obtain estimated values for the future window $\hat{X}_f$ and its reconstruction loss $\mathcal{L}_{3}$:

\begin{equation}
    \begin{aligned}
            \hat{X}_f &= \mathcal{D}_{(\mu_h,\sigma_h)}\{\hat{X}_{f,trend} + \hat{X}_{f,season}\}, \\
        \mathcal{L}_{3} &= \|{X}_f - \hat{X}_f \|^2,
    \end{aligned}
    \label{eq:pred_X}
\end{equation}

\noindent and the overall optimization function \( \mathcal{L}_{joint} \) for the joint training phase is as follows:

\begin{equation}
    \begin{aligned}
         \mathcal{L}_{joint} = \mathcal{L}_{3} + \gamma \mathcal{L}_{trend} , 
    \end{aligned}
\label{eq:tokenizer}
\end{equation}

\noindent where $\gamma$ regulates the balance between the trend representation task and the final forecasting task.

\subsubsection{Training Stage \uppercase\expandafter{\romannumeral2}}

To enable the diffusion-based token predictor to learn how to iteratively restore masked tokens, we design a denoising training procedure. As shown in the right of Figure \ref{fig:tokenizer}, given the encoded historical trend tokens \(T_h\) and future trend tokens \(T_f\), we randomly replace a subset of tokens in \(T_f\) with [mask] according to a noise scheduler, resulting in partially masked future tokens $\tilde{T_f}$. We then construct the model input as $\tilde{T_t}=Concat(T_h, \tilde{T_f})$ and the supervision signal as \(T_t=Concat(T_h, T_f)\). The model is optimized by a cross-entropy loss to recover the masked tokens:  

\vspace{-0.05in}
\begin{equation}
    \mathcal{L}_{diffusion} = -\frac{1}{\sum_{i=1}^{N} \mathbb{I}_{\text{mask}}(i)} \sum_{i=1}^{N} \mathbb{I}_{\text{mask}}(i) \sum_{j=1}^{C} y_{ij} \log(\hat{y}_{ij}) ,
\end{equation}


\noindent where \(N\) is the total number of tokens in \(T_t\), \(y_{ij}\) denotes the true label of the \(i\)-th token at class \(j\), and \(\mathbb{I}_{\text{mask}}\) is an indicator function that equals 1 when the token is masked and 0 otherwise. In this way, only masked positions contribute to the loss, guiding the model to focus on denoising. This training scheme equips the diffusion predictor with the ability to progressively refine partially masked sequence into complete and coherent future token representations.




\subsection{Prediction Process of OneCast} When a historical window comes, OneCast first normalizes and then 
decomposes it into seasonal and trend component using moving
average. For the seasonal part, OneCast leverages seasonal encoder to predict the various coefficients of basis functions and then construct the future season component. As for the trend part, OneCast employs the semantic tokenizer to encode it into a discrete token sequence, which then serves as information to guide the diffusion-based token predictor in generating future tokens; finally, it utilizes future decoder to decode these future tokens into the future trend series.  Ultimately, the predicted future seasonal and trend  parts are combined and denormalized to obtain final predictions.



\section{Experiments}
\label{sec:exp}

\subsection{Experimental Setup}


\subsubsection{Datasets and Baselines}
To demonstrate the effectiveness of OneCast in cross-domain prediction, we conduct extensive experiments on long-term multi-variate time series forecasting tasks under 9 diverse range of real-world datasets from various domains, including Traffic \cite{trafficdata}, ETTh2 \cite{zhou2021informer}, ETTm2 \cite{zhou2021informer}, Weather \cite{weatherdata}, CzeLan \cite{poyatos2016sapfluxnet}, FRED-MD \cite{mccracken2016fred}, NYSE \cite{panagopoulos2021transfer}, Covid-19 \cite{panagopoulos2021transfer} and Wike2000 \cite{gasthaus2019probabilistic}. On these datasets, we conduct comparative analysis against various advanced baselines, including three cross-domain methods: representation-alignment based UniTime \cite{liu2024unitime} and TimesFM \cite{das2024decoder}, and discretization-based TOTEM \cite{talukder2024totem}; and six in-domain methods from different architectures: Transformer-based PatchTST \cite{nie2022time}, Autoformer \cite{wu2021autoformer}, Fedformer \cite{zhou2022fedformer}; MLP-based DLinear \cite{zeng2023transformers}; CNN-based MICN \cite{wang2023micn}; and Legendre Memory-based FiLM \cite{zhou2022film}. Detailed descriptions are shown in Appendix \ref{appendixA}.

\subsubsection{Implementation Details.}  Refer to the common benchmark practices \cite{qiu2024tfb, wang2024deep}, we carefully set the target prediction lengths accordingly. Specifically, for the five large-scale datasets with observed time points more than 10,000, the historical input window is fixed at 96, with forecasting horizons of \{24, 48, 96, 192\}; for the four small-scale datasets, the input window is fixed at 36, with horizons of \{24, 36, 48, 60\}. All experiments are implemented in PyTorch \cite{paszke2017automatic} and run on a single NVIDIA A800 80GB GPU, except UniTime which uses 4 GPUs. Baselines are mainly from TSlib \cite{wang2024tssurvey}, with others from official libraries. We evaluate all experiments using mean squared error (MSE) and mean absolute error (MAE) :
\begin{equation}
    \begin{aligned}
        \text{MSE} &= \frac{1}{n} \sum_{i=1}^{n} (y_i - \hat{y}_i)^2, \\
        \text{MAE} &= \frac{1}{n} \sum_{i=1}^{n} |y_i - \hat{y}_i|,
    \end{aligned}
\end{equation}

\noindent where \( y_i \) is the actual value, \( \hat{y}_i \) is the predicted value, and \( n \) is the number of observations. Each experiment is repeated three times and averaged, and lower values indicate better performance. More implementation details of OneCast can be found in Appendix \ref{appendixB}.

\begin{table}[ht]
    \centering
    \caption{Full dataset descriptions.}
    \begin{tabular}{@{}ccccc@{}}
        \toprule
         Dataset & Domain & Frequency & Samples & Variables \\ 
        \midrule
        Traffic & Traffic & 1 hour & 17,544 & 862  \\ 
        ETTh2 & Electricity & 1 hour & 14,400 & 7  \\ 
        ETTm2 & Electricity & 15 mins & 57,600 & 7 \\ 
        Weather & Environment & 10 mins & 52,696 & 21  \\ 
        CzeLan & Nature & 30 mins & 19,934 & 11  \\ 
        FRED-MD & Economic & 1 month & 728 & 107   \\ 
        NYSE & Stock & 1 day & 1,244 & 5  \\ 
        Covid-19 & Health & 1 day & 1,392 & 948 \\ 
        Wike2000 & Web & 1 day & 792 & 2,000 \\ 
        \bottomrule
    \end{tabular}
    \label{appendix_tab:dataset-desc}
\end{table}

\begin{table*}[h]
\centering
\small
\caption{Performance comparison of OneCast and baseline models, highlighting the best values in \textbf{bold} and the second-best values {\ul underlined}. All results are averaged MSE$\downarrow$ and MAE$\downarrow$ from four different predicted windows.}
\setlength{\tabcolsep}{2.5pt}
\resizebox{\textwidth}{!}{
\begin{tabular}{@{}c|cccccccc|cccccccccccc@{}}
\toprule
Type & \multicolumn{8}{c|}{Models Trained Across-Domain} & \multicolumn{12}{c}{Models Trained In-Domain} \\ 
\cmidrule(l){2-21} 
Methods & \multicolumn{2}{c}{OneCast} & \multicolumn{2}{c}{UniTime} & 
\multicolumn{2}{c}{TOTEM} & \multicolumn{2}{c|}{TimesFM} &
\multicolumn{2}{c}{PatchTST} & \multicolumn{2}{c}{FEDformer} & \multicolumn{2}{c}{Autoformer} & 
\multicolumn{2}{c}{FILM} & \multicolumn{2}{c}{DLinear} &
\multicolumn{2}{c}{MICN} \\
Metric & MSE & MAE & MSE & MAE & MSE & MAE & MSE & MAE & MSE & MAE & MSE & MAE & MSE & MAE & MSE & MAE & MSE & MAE & MSE & MAE \\
\midrule 

Traffic & \textbf{0.492} & 
\underline{0.330} & 0.543 & 0.363 & 0.617 & 0.350 & 0.543 & 0.350 & 0.562 & 0.368 & 0.574 & 0.363 & 0.594 & 0.377 & 0.604 & 0.393 & 0.615 & \textbf{0.307} & \underline{0.511} & 0.370 \\
Weather & \textbf{0.173} & \textbf{0.213} & 0.177 & 0.218 & 0.186 & 0.232 & 0.189 & 0.220 & \underline{0.174} & \underline{0.214} & 0.227 & 0.304 & 0.237 & 0.309 & 0.179 & 0.236 & 0.178 & 0.234 & \textbf{0.173} & 0.225 \\
CzeLan & \textbf{0.206} & \textbf{0.261} & 0.239 & 0.287 & 0.228 & 0.287 & 0.221 & 0.279 & \underline{0.212} & \underline{0.266} & 0.247 & 0.323 & 0.575 & 0.516 & 0.287 & 0.371 & 0.296 & 0.362 & 0.276 & 0.337 \\
ETTh2 & \underline{0.279} & \underline{0.336} & 0.280 & 0.338 & 0.323 & 0.366 & 0.280 & 0.342 & \textbf{0.269} & \textbf{0.331} & 0.323 & 0.379 & 0.348 & 0.398 & 0.310 & 0.362 & 0.304 & 0.387 & 0.331 & 0.349 \\
ETTm2 & 0.185 & 0.265 & \underline{0.177} & 0.264 & 0.192 & 0.276 & 0.182 & 0.277 & \textbf{0.167} & \textbf{0.252} & 0.191 & 0.280 & 0.208 & 0.295 & 0.172 & 0.278 & 0.185 & 0.273 & 0.184 & \underline{0.256} \\
FRED-MD & \textbf{69.70} & \textbf{1.270} & 92.283 & 1.650 & \underline{72.05} & 1.579 & 75.89 & 1.348 & 87.66 & 1.602 & 116.5 & 2.023 & 116.7 & 2.088 & 118.7 & 2.069 & 122.2 & 2.442 & 144.9 & 1.965 \\
NYSE & \textbf{0.432} & \textbf{0.417} & 0.488 & 0.467 & 0.512 & 0.474 & \underline{0.450} & \underline{0.437} & 0.544 & 0.482 & 0.518 & 0.488 & 0.676 & 0.573 & 0.684 & 0.754 & 1.004 & 0.735 & 0.928 & 0.566 \\
Covid-19 & \textbf{1.533} & \textbf{0.059} & 1.710 & \textbf{0.059} & 2.320 & 0.088 & 1.645 & \underline{0.062} & \underline{1.644} & \underline{0.062} & 2.579 & 0.209 & 2.617 & 0.266 & 2.132 & 0.455 & 28.588 & 0.688 & 72.588 & 0.072 \\
Wike2000 & \textbf{557.5} & \textbf{1.176} & 630.7 & \underline{1.244} & 678.9 & 1.316 & \underline{582.3} & 1.311 & 584.8 & 1.250 & 718.0 & 3.279 & 722.5 & 3.303 & 1178 & 1.398 & 632.3 & 1.521 & 645.7 & 1.558 \\

\bottomrule
\end{tabular}
}
\label{tab:main_table}
\end{table*}

\subsection{Forecasting Performance}


Table \ref{tab:main_table} presents the results of the long-term forecasting experiments. The table is partitioned by two vertical lines. Models on the left are trained jointly across cross-domain datasets, while those on the right undergo separate training for each dataset. As shown in table, our OneCast achieves state-of-the-art results in 13 out of 18 entries when compared to all baselines, demonstrating strong competitive performance across both cross-domain and in-domain models. These outcomes validate the effectiveness of our model in handling diverse data characteristics across domains.

Notably, OneCast achieves outstanding results on datasets with distinct temporal characteristics, such as Traffic and Wike2000. This advantage arises from its structural decomposition and parallel processing strategy: the seasonal projection module explicitly captures periodic signals in Traffic data, while the discrete token-based diffusion branch effectively models sudden patterns in Wike2000. In contrast, PatchTST also performs competitively, likely due to its patch-based channel-independent  modeling that captures temporal semantics while reducing inter-channel complexity. Nevertheless, OneCast shows relatively poor performance on ETTh2 and ETTm2. It is hypothesized that both of them are electricity load datasets whose fluctuations depend not only on periodicity but also on abrupt consumption changes and strong local nonlinearities. OneCast may prioritize transferring general temporal regularities, thereby showing limited capacity to capture fine-grained, domain-specific variations.

\subsection{Ablation Study}

\subsubsection{Effective of Cross-domain Training} 
\begin{figure}[htbp]
    \centering
    \includegraphics[width=\linewidth]{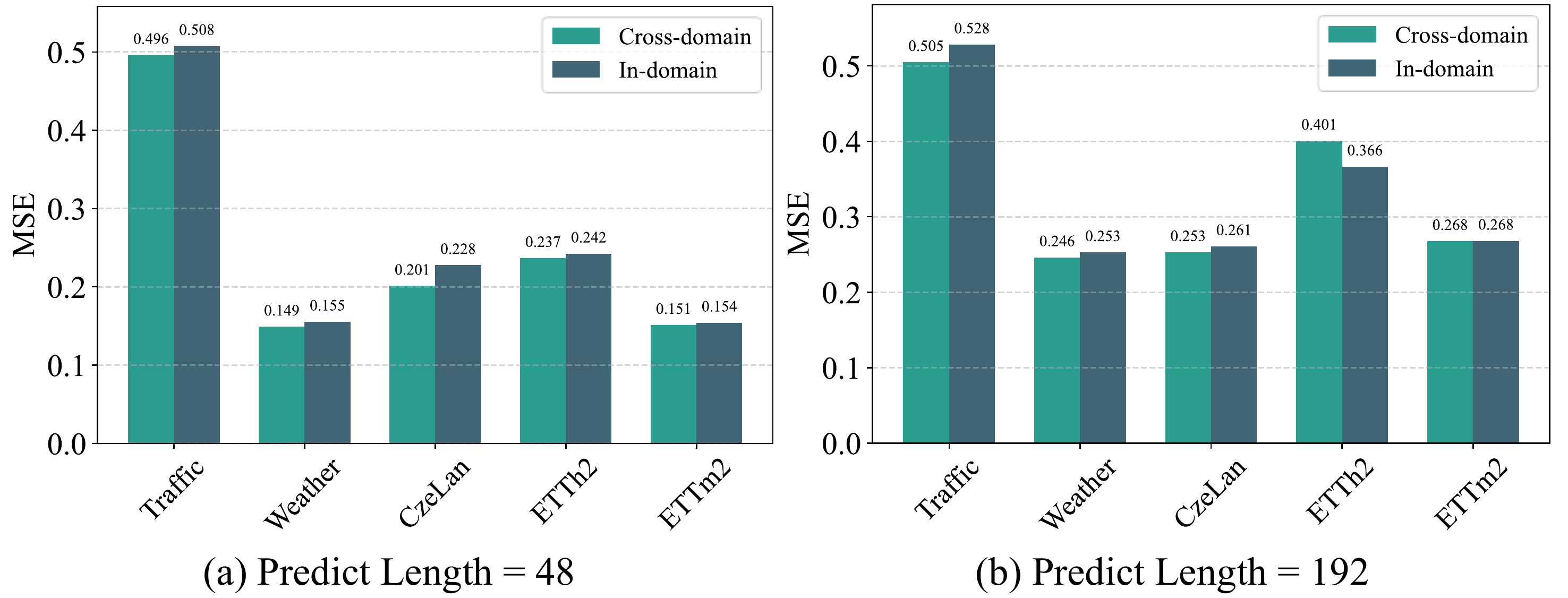}
    \caption{The forecasting performance of OneCast between cross-domain and in-domain training, with different prediction length of 48 and 192.}
    \label{fig:cross}
\end{figure}

To verify the effectiveness of cross-domain training, we compare the forecasting performance of OneCast trained in-domain. As shown in Figure \ref{fig:cross}, OneCast trained across domain achieves better performance than in-domain training in most cases, further confirming the advantages of cross-domain training in suitable scenarios, demonstrating its strong generalization ability. Notably, however, the gains from cross-domain training diminish as the prediction horizon increases; in particular, ETTh2 exhibits negative transfer at the 192-step horizon. This plausibly due to weaker alignment of temporal patterns between different datasets under the larger predict window, which reduces the benefits of cross-domain transfer.

\definecolor{darkgreen}{RGB}{0, 100, 0}
\definecolor{darkred}{RGB}{100, 0, 0}

\setlength{\heavyrulewidth}{0.12em}

\subsubsection{Effective of Dual-decoder Strategy} 

To vertify that the dual-decoder training strategy can excels in reconstructing future trends under cases of distribution shift, we compare it with the single-decoder training strategy. We first define the Average Mean Absolute Difference (AMAD) as follows: 

\begin{equation}
    AMAD(X, \hat{X}) = \frac {1}{N} \sum_{i=1}^N \left| \mu_{x_i} - \mu_{\hat {x}_i} \right|, 
\end{equation}

\noindent where $N$ is the total number of samples in dataset and $\mu$ refers to the mean of sample,  with $X$ denoting the ground truth and $\hat{X}$ represents the prediction results. The smaller the AMAD, the stronger the model's ability to fit the future mean. Table \ref{tab:dual_vs_single} shows the difference between $AMAD_{dual}$ and $AMAD_{single}$, where most values are less than 0, indicating that the dual-decoder training strategy yields consistently lower AMAD on most datasets across all horizons (except a small positive deltas persist on Traffic and FRED-MD), demonstrating the stronger ability it holds in future distribution fitting. 

\begin{figure}[htbp]
    \centering
    \includegraphics[width=\linewidth]{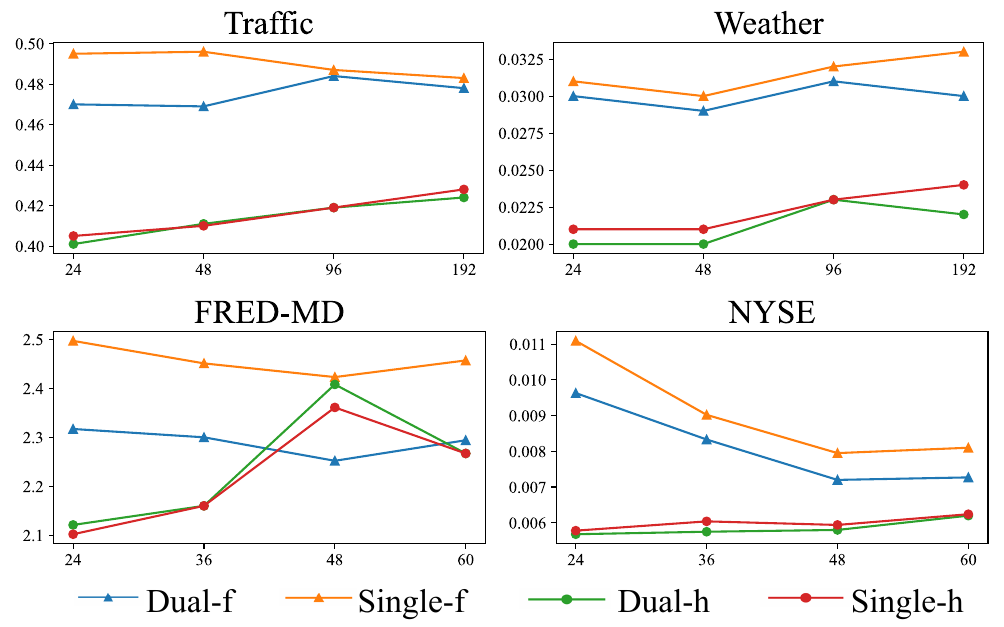}
    \caption{Comparison of dual- and single-decoder training strategies. "Dual-h" and "Dual-f" are reconstruct errors of historical and future windows under dual-decoder strategy; "Single-h" and "Single-f" denote those under the single one.}
    \label{fig:dual_dec}
\end{figure}

Building on this, we further compare the two strategies in terms of sequence compression capability and future window token decoding ability. As illustrated in Figure \ref{fig:dual_dec}, the tokenizer trained with a dual-decoder strategy exhibits a reconstruction error for the historical window that is roughly equivalent to the single-decoder trained. However, for future windows, the dual-decoder strategy significantly outperforms the latter. These findings suggest that the dual-decoder training strategy not only maintains the compression capability for sequential data but also significantly enhances the tokenizer's ability to decode future windows using historical statistical information.


\begin{figure*}[htbp]
    \centering
    \includegraphics[width=\textwidth]{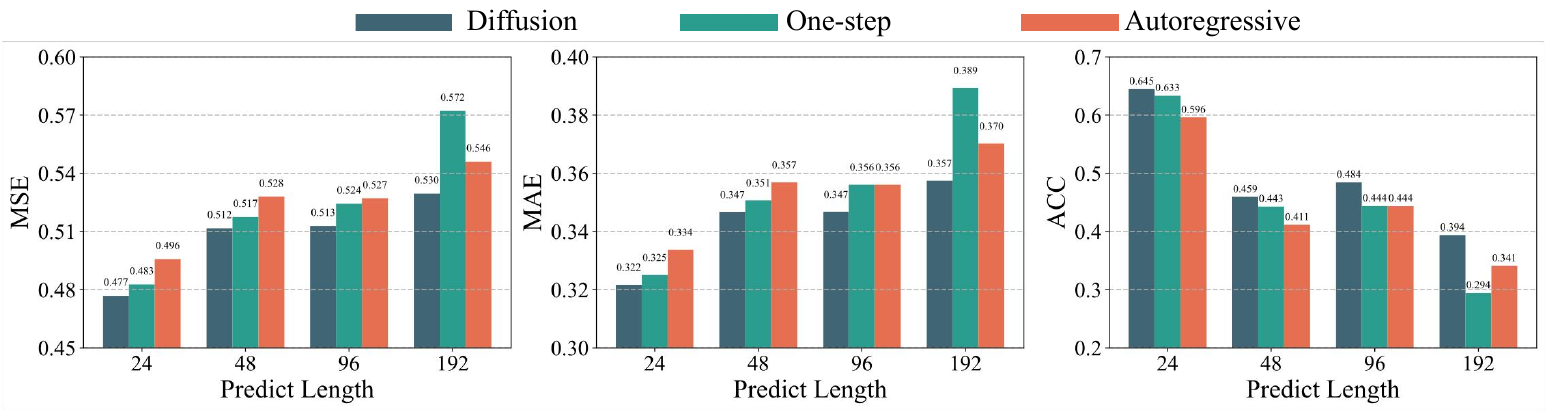}

    \vspace{-0.1in}
    \caption{Comparison of three strategies for token predictor on Traffic: diffusion, one-step generation, and autoregressive, evaluated by the MSE and MAE between decoded future series and ground truth, as well as the accuracy of predicted tokens.}
    \label{fig:diffusion}
\end{figure*}

\begin{table*}[h]
    \centering
    \large
    \caption{Comparison of different mask strategies with future window length 192, where $t \in [0,1]$ for all methods.}
    \vspace{-0.1in}
    \begin{tabular}{@{}c|c|cc|cc|cc|cc|cc|cc@{}}
        \toprule
        \multirow{2}{*}{Method} & \multirow{2}{*}{Formula} & \multicolumn{2}{c|}{ETTh2} & \multicolumn{2}{c|}{ETTm2} & \multicolumn{2}{c|}{Weather} & \multicolumn{2}{c|}{Traffic} &
        \multicolumn{2}{c|}{CzeLan} & \multicolumn{2}{c}{Avg.} \\
        \cmidrule(lr){3-14}
         &  & mse & mae & mse & mae & mse & mae & mse & mae & mse & mae & mse & mae \\
        \midrule
        cosine & $\cos\!\left(t \cdot \tfrac{\pi}{2}\right)$ & 0.401 & 0.408 & 0.268 & 0.318 & \textbf{0.246} & 0.273 & 0.505 & 0.338 & 0.253 & 0.285 & 0.335 & \textbf{0.324} \\ 
        linear & $1 - t$ & \textbf{0.400} & \textbf{0.407} & \textbf{0.265} & 0.318 & 0.247 & 0.274 & 0.503 & \textbf{0.337} & 0.251 & 0.283 & \textbf{0.333} & \textbf{0.324} \\ 
        power & $1 - t^2$ & 0.407 & 0.410 & \textbf{0.265} & \textbf{0.317} & 0.247 & \textbf{0.272} & 0.505 & 0.339 & 0.253 & 0.286 & 0.335 & 0.325 \\ 
        sigmoid & $\dfrac{\sigma(t)-\sigma(0)}{\sigma(1)-\sigma(0)}$ & 0.410 & 0.412 & \textbf{0.265} & \textbf{0.317} & 0.248 & 0.274 & \textbf{0.502} & \textbf{0.337} & \textbf{0.247} & \textbf{0.281} & 0.334 & \textbf{0.324} \\ 
        \bottomrule

    \end{tabular}
    \label{appendix_tab:masking_results}
\end{table*}

\begin{table}[htbp]
\centering
\footnotesize 

\caption{Values of \textbf{$AMAD_{dual} - AMAD_{single}$} , where values < 0 indicate the dual decoder strategy better fits future means.}

\resizebox{0.9\linewidth}{!}{
\begin{tabular}{@{}c|cccc@{}}
\toprule
\textbf{Pred. Len.} & 24 & 48 & 96 & 192 \\ \midrule 
Traffic & \textcolor{darkred}{+0.001} & \textcolor{darkred}{+0.085} & \textcolor{darkred}{+0.046} & \textcolor{darkred}{+0.018} \\
Weather & {-0.029} & {-0.020} & {-0.011} & {-0.008} \\
CzeLan & {-0.057} & {-0.080} & {-0.067} & {-0.070} \\
ETTh2 & {-0.053} & {-0.025} & {-0.024} & {-0.007} \\
ETTm2 & {-0.053} & {-0.025} & {-0.024} & {-0.007} \\ \midrule
\textbf{Pred. Len.} &  24 & 36 & 48 & 60 \\ \midrule
Wike2000 & {-0.011} & {-0.011} & {-0.009} & {-0.031} \\
FRED-MD & {-0.076} & {-0.022} & {-0.083} & {-0.115} \\
Covid-19 & \textcolor{darkred}{+0.009} & \textcolor{darkred}{+0.009} & \textcolor{darkred}{+0.014} & \textcolor{darkred}{+0.012} \\
NYSE & {-0.079} & {-0.126} & {-0.108} & {-0.098} \\ \bottomrule
\end{tabular}}

\label{tab:dual_vs_single}
\end{table}

\subsubsection{Effective of Discrete Diffusion Strategy}

To validate the effectiveness of the discrete diffusion training paradigm, we compare one-step (generate all tokens at once) and autoregressive token generation on the Traffic dataset, which features large data volume and multiple channels. As shown in Figure \ref{fig:diffusion}, as the prediction window grows, the number of tokens increases linearly and the one-step approach degrades markedly, falling below both autoregressive and diffusion variants at the 192-step horizon. For shorter horizons (24 and 48), however, one-step generation remains competitive or superior to autoregression, likely because it avoids cumulative error. This pattern reflects that one-step generation lacks intermediate corrective feedback and can hit capacity bottlenecks when producing long sequences in a single pass. By contrast, diffusion iteratively refines multiple tokens with confidence-aware updates, stabilizing long-range dependencies, and consequently achieving the best accuracy across all four settings.

\subsubsection{Varying Mask Scheduler of Diffusion}
The masking strategy is a key hyperparameter in mask diffusion \cite{he2022diffusionbert,nie2025large}, as it controls how the visible token ratio evolves during the denoising trajectory. We compare four representative schedulers—cosine, linear, power, and sigmoid—each corresponding to a distinct decay pattern of the masking ratio. The mathematical formulations and results of them are summarized in Table~\ref{appendix_tab:masking_results}. The overall average performance remains largely consistent across different masking strategies, indicating that the diffusion framework is relatively robust to the choice of scheduler. However, distinct strengths emerge across specific datasets. For instance, the linear scheduler achieves the lowest MSE on ETTh2, showing an improvement of approximately 2.5\% over the sigmoid scheduler. In contrast, the sigmoid scheduler performs best on CzeLan, reducing MSE by about 2.4\% compared with the other schedulers. These findings suggest that while diffusion models are generally insensitive to the masking function, tailoring the scheduler to the dataset’s temporal characteristics can yield further performance gains.



\subsection{In-depth Analysis of OneCast}

To further illustrate the ability of OneCast in addressing cross-domain challenges, we analyze its capabilities by examining common issues associated with cross-domain scenarios.


\begin{table*}[htbp]
\centering

\caption{The comparison involves the tokenizer's reconstruction error (Reconst.) and OneCast's final prediction error (Final Pred.), both measured by MSE. Results are averaged over four prediction lengths, and "Rate." is the ratio of reconstruction to final prediction MSE; a smaller rate suggests less impact of reconstruction error on final outcomes.}

\resizebox{0.95\textwidth}{!}{
\begin{tabular}{@{}c|cccccccccc@{}}

\toprule
Dataset & Traffic & Weather & CzeLan & ETTh2 & ETTm2 & FRED-MD & NYSE & Covid-19 & Wike2000 \\ \midrule
Reconst. & 0.042 & 0.010 & 0.010 & 0.016 & 0.007 & 7.569 & 0.011 & 0.618 & 263.662 \\
Final Pred. & 0.492 & 0.173 & 0.206 & 0.278 & 0.184 & 69.701 & 0.432 & 1.533 & 557.510 \\ \midrule
Rate. & 8.53\% & 5.78\% & 4.85\% & 5.73\% & 3.77\% & 10.41\% & 2.54\% & 40.31\% & 47.29\% \\ \bottomrule
\end{tabular}}

\label{tab:reconst_info_loss}
\end{table*}

\begin{figure}[htbp]
    \centering
    \subfigure{
        \includegraphics[width=0.47\linewidth]{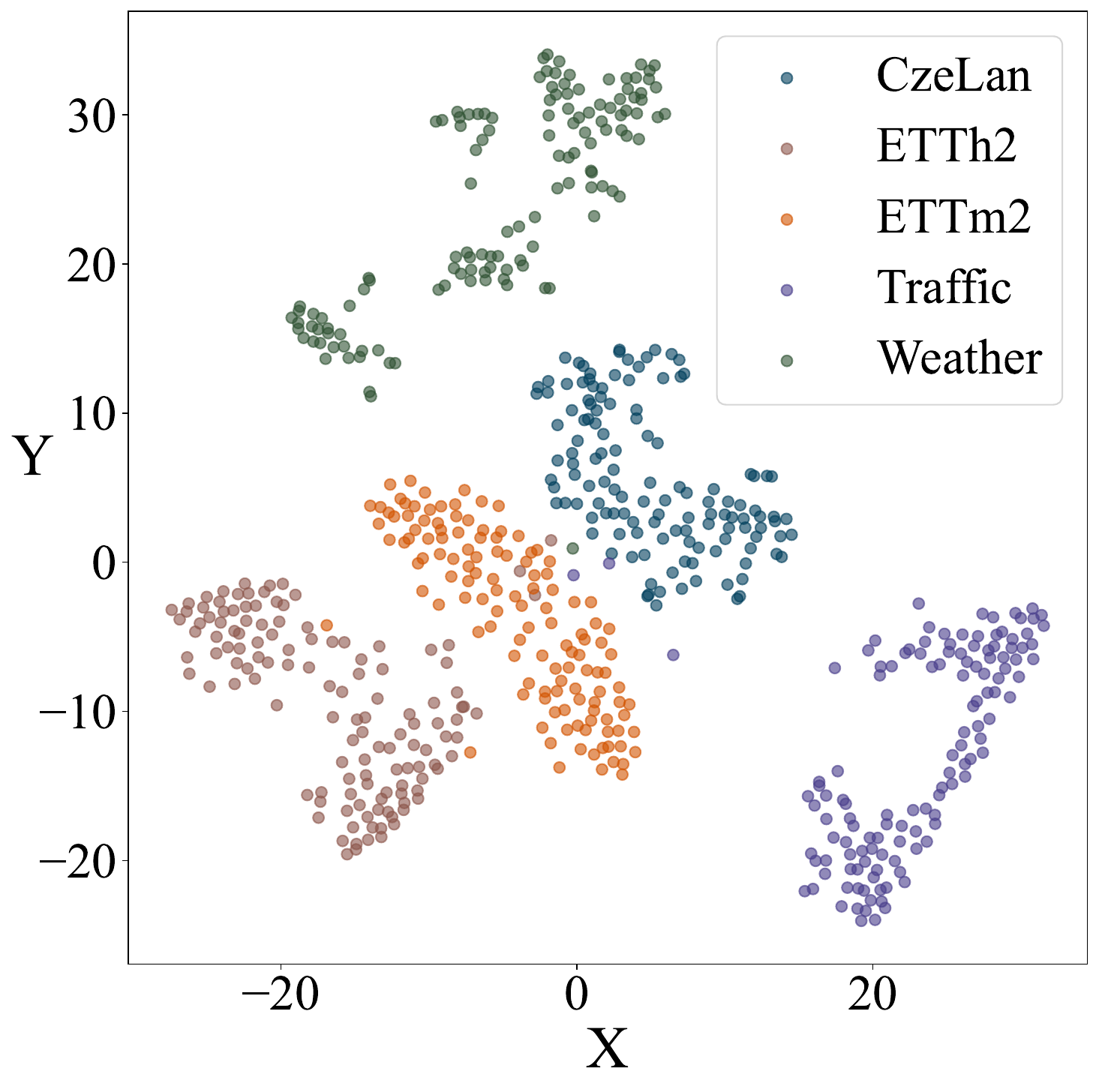}
        \label{fig:tsne_codebook}
    }
    \subfigure{
        \includegraphics[width=0.47\linewidth]{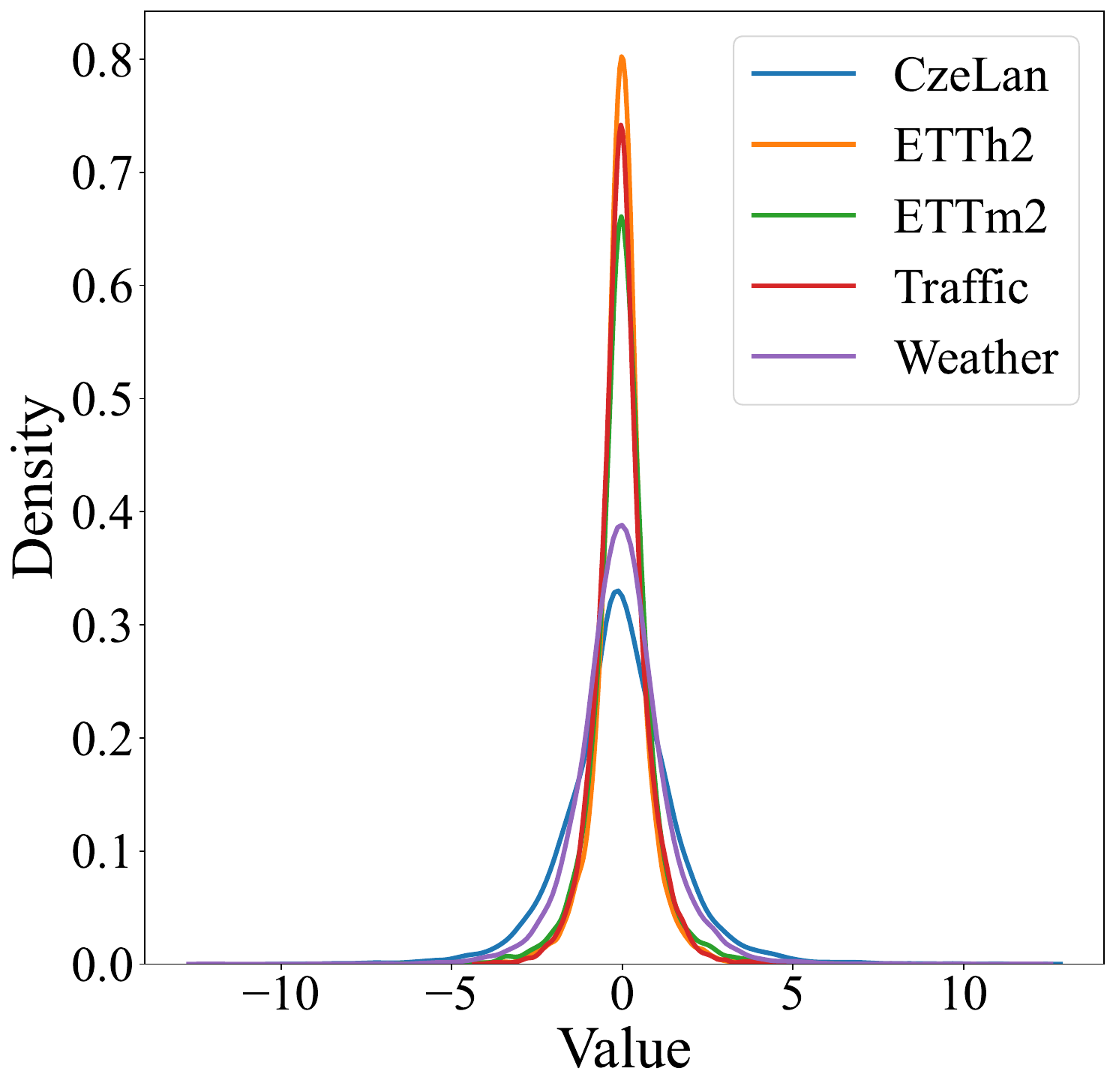}
        \label{fig:kde_codebook}
    }

    \vspace{-0.2in}
    \caption{Visualization of codebook vector from different domains. Left: t-SNE of codebook vectors; Right: probability density distribution of the codebook values.}
    \label{fig:codebook}
\end{figure}

\subsubsection{Ability on Solving Domain Confusion}
Domain confusion is a key challenge in cross-domain models, where difficulty in identifying the domain origin of data undermines prediction accuracy \cite{liu2024unitime}.  To mitigate potential domain confusion, we visualize the distribution of codebook vectors across domains. Figure \ref{fig:tsne_codebook} shows that t-SNE visualization reveals distinct spatial distributions of codebook vectors across domains. To rule out the impact of numerical scale differences, we calculate the probability densities of the values of codebook vectors from different domains. Results in Figure \ref{fig:kde_codebook} confirm that numerical scales are broadly consistent across domains. This implies that cross-domain token-indexed latent vectors have discriminative distributions in a uniform scale space, facilitating discrimination by downstream token predictors. Thus, the model can effectively distinguish cross-domain data, thereby avoiding domain confusion and supporting unified cross-domain training.

\subsubsection{Low Information Loss in Unified Representation}

When concerning the inconsistency in feature dimensions across domains, OneCast resolves this by uniformly encoding cross-domain sequence of varying dimensions into discrete tokens. To demonstrate the minimal information loss in this discretization process, We compare the series reconstruction MSE with the final prediction MSE and report the ratio of reconstruction to prediction error. Table \ref{tab:reconst_info_loss} shows that the median rate is 5.78\% across all of the nine datasets, and most are under 10\%, demonstrating the remarkable efficacy of our tokenizer in compressing temporal information with minimal loss. The two exceptions, Covid-19 (40.31\%) and Wike2000 (47.29\%), exhibit relatively higher reconstruction errors, primarily attributed to the inherent challenges of training a well-generalized tokenizer on limited data with high channel dimensionality.



\subsubsection{Low Token Consumption in Series Encoding}

OneCast reduces token consumption during series encoding through a cross-channel encoding scheme. For a more detailed analysis, we compare three mainstream tokenization schemes using their representative methods: continuous patching-based TimesFM, value discretization-based Chronus, and text representation-based PromptCast. We uniformly set the sequence length $L$ to 96, patch size $P$ to 16, and the number of text tokens $k$ to 3 for representing each data point in PromptCast. The seasonal vocabulary size $M$ in OneCast is set to 437, with more details are provided in Appendix \ref{appendix:tokenizer}. Table \ref{tab:token_consumption} presents the number of tokens required to encode sequence under different datasets with varying feature counts $C$ (i.e., channels). 

\begin{table}[htbp]
    \centering

    \caption{Token counts required for encoding sequence with different methods, where \# indicates the number of channels.}

    \resizebox{\linewidth}{!}{
    \begin{tabular}{@{}c|ccccc@{}}
        \toprule
        \multirow{2}{*}{\textbf{Methods}} & \multirow{2}{*}{\textbf{Complexity}} & \textbf{CzeLan} & \textbf{FRED-MD} & \textbf{Traffic} & \textbf{Wike2000} \\ 
        & & \#11  & \#107 & \#862 & \#2000 \\ \midrule 
        TimesFM & $(\lceil L / P \rceil ) * C$ & 66 & 642 & 5,172 & 12,000 \\
        Chronus & $L * C$ & 1,056 & 10,272 & 82,752 & 192,000 \\
        PromptCast & $k * L * C$ & 3,168 & 30,816 & 248,256 & 576,000 \\ \midrule
        OneCast & $(\lceil L / P \rceil ) + M$ & 443 & 443 & 443 & 443 \\ \bottomrule
    \end{tabular}}

    \label{tab:token_consumption}
\end{table}

As shown in table, OneCast exhibits a distinct efficiency advantage compared to other models, which token count remaining independent of feature counts. In high-dimensional scenarios with 862 channels, the token count of OneCast is merely 8.5\% of that of TimesFM, 0.53\% of that of Chronus, and 0.17\% of that of PromptCast. This enables OneCast to significantly reduce resource consumption when processing large-scale time series data with long sequence and large channel counts, thereby providing critical support for efficient deployment in practical scenarios.


\section{Conclusion}

In this work, we proposed OneCast, a structured framework for cross-domain time series forecasting. By explicitly decomposing seasonal and trend components, OneCast performs specialized modeling that enhances both generalization and interpretability. The seasonal part is reconstructed through lightweight periodic basis functions, enabling efficient and interpretable recovery of recurring patterns. For the trend part, we designed a unified tokenization module with a dual-decoder training strategy, which jointly balances representation compression and predictive capability. A discrete diffusion-based token generator further enables confidence-aware, parallel forecasting, alleviating error accumulation inherent in autoregressive designs. We conducted extensive experiments across 9 real-world datasets, demonstrating the exceptional cross-domain forecasting capability of OneCast. We hope this work could offer a new perspective for cross-domain sequence modeling.


\clearpage

\bibliographystyle{ACM-Reference-Format}
\bibliography{OneCast}


\appendix

\input{appendix/appendix_A}

\input{appendix/appendix_B}

\input{appendix/appendix_C}

\input{appendix/appendix_D}

\input{appendix/appendix_E}

\input{appendix/appendix_F}

\input{appendix/appendix_G}

\end{document}

%% file: appendix/appendix_A.tex
\section{Experimental Setup}
\label{appendixA}

\subsection{Dataset Descriptions}
To evaluate the cross-domain predictive ability of OneCast, we conduct experiments on nine datasets spanning traffic, electricity, environment, nature, economics, stock markets, health, and web analytics. As shown in Table \ref{appendix_tab:dataset-desc}, they cover diverse sampling rates, variable dimensions, and scales, including five large datasets with over 10,000 samples and four smaller ones with about 1,000, enabling assessment under both rich and limited data scenarios.



\textbf{Traffic\cite{trafficdata}} The hourly road occupancy data for San Francisco Bay Area highways collected by 862 sensors along the highways, covering the time period from January 2015 to December 2016.

\textbf{ETT\cite{zhou2021informer}} This dataset comprises four subsets (ETTh1, ETTh2, ETTm1, ETTm2), each capturing the load characteristics of seven types of oil and power transformers from July 2016 to July 2018. Among these, ETTh1 and ETTm1 are 1-hour-level subsets, while ETTh2 and ETTm2 are 15-minute-level subsets. We conduct our experiments using the hour-level ETTh2 and the minute-level ETTm2.

\textbf{Weather\cite{weatherdata}} The meteorological time series featuring 21 indicators, which was collected every 10 minutes throughout 2020 by the weather station at Max Planck Institute for Biogeochemistry.
 
\textbf{CzeLan\cite{poyatos2016sapfluxnet}} Sap flow measurements and environmental variables collected from the Sapflux project.

\textbf{FRED-MD\cite{mccracken2016fred}} The macroeconomic dataset provided by the Federal Reserve Bank of St. Louis, comprises over 100 monthly time series that describe the condition of the U.S. economy.

\textbf{NYSE\cite{feng2019temporal}} The dataset records opening price, closing price, trading volume, lowest price, and highest price comes from stocks traded on the New York Stock Exchange (NYSE) between January 2013, and December 2017.

\textbf{Covid-19\cite{panagopoulos2021transfer}} The dataset includes the number of COVID-19 cases in different regions of the four considered European countries: Italy, Spain, France, and England, combining this case with the aggregated mobility data provided by Facebook. 

\textbf{Wike2000\cite{gasthaus2019probabilistic}} The dataset records daily page views of 2000 Wikipedia pages.

%% file: appendix/appendix_B.tex
\subsection{Implement Details of OneCast}
\label{appendixB}





In the training process of OneCast, the dataset partitioning consistently aligns with the baselines, following the proportions provided in Time Series Forecasting Benchmark \cite{qiu2024tfb}.

\subsubsection{Training Stage \uppercase\expandafter{\romannumeral1}: Joint Optimization}
\label{appendix:tokenizer}

We adopt AdamW \cite{loshchilov2017decoupled} with learning rate $5\times10^{-4}$, weight decay $1\times10^{-5}$, and LambdaLR decay (0.99 every 300 steps) as optimizer, training up to 25 epochs. The best model is chosen by validation $\mathcal{L}_{joint}$.

For the seasonal components, we use a 2-layer MLP (hidden=64) to predict future weights ${v_{j,c}^i}$, initialized with 437 commonly used basis functions accordingly \cite{darlow2024dam}. For the trend tokenizer, we adopt a 3-block 1D Conv VQ-VAE (kernel=3) as encoder–decoder, with a feature extractor that partitions every 16 time points (kernel=8) into 2 tokens. The codebook contains 128 vectors of dimension 64, and the encoder is trained with $\beta=0.25$.

\subsubsection{Training Stage \uppercase\expandafter{\romannumeral2}: Diffusion-based Token Predictor}
\label{Token Predictor}

We train the token predictor with a discrete diffusion method \cite{xie2024show,he2022diffusionbert}, using a 2-layer decoder-only transformer (hidden=128) as our backbone. The \textit{[mask]} embedding is initialized as the mean of temporal token embeddings to reduce distribution gap. A cosine scheduler controls the mask probability:

\begin{equation}
    p(t) = \cos\left(\frac{\pi t}{2}\right), \quad t\sim U[0,1),
\end{equation}



\noindent where normalized step $t \in [0, 1) $ is randomly selected, and $p(t) \in (0,1]$ refers to the mask probability that would be used for noise addition. In the inference phase, we set the number of inference steps to 4, ensuring that each prediction of the future token sequence undergoes 4 rounds when denoising. Detailed training and inference algorithms of token predictor are shown in Algorithm \ref{appendix_alg:train} and Algorithm \ref{appendix_alg:infer}, respectively.

\begin{algorithm*}[htbp]
  \small
  \caption{Training Process of Token Predictor}\label{algo:OneCast}
  \begin{algorithmic}[1]
    \Require  
   The discretized historical tokens \( T_h \in \mathbb{N}^{B \times L_h}\) and future token \( T_f \in \mathbb{N}^{B \times L_f} \) with batch size $B$, transformed time series vocabulary $\hat{E} = E \times M \in \mathbb{R}^{K \times D} $,   mask embedding $E_{mask} \in \mathbb{R}^{1 \times D}$, mask scheduler MS , and decode-only transformer-based backbone TRM.

    \State $t \sim U[0,1)^B$
    \Comment Randomly sample the normalized noisy step $t \in [0, 1)^{B}$
    
    \State $p_{mask} = \text{MS}(t)$
    \Comment{Initialize the mask probability $p_{mask} \in [0,1)^{B}$ by MS and $t$}

    \State $\tilde{T_f} = T_f$
    \Comment{Initialize $\tilde{T_f}$ as a copy of $T_f$}
    
    \For{each batch $i \in \{1, \ldots, B\}$}
      \For{each token $j \in \{1, \ldots, L_f\}$}
        \State $r \sim U[0,1) $
        \Comment{Sample a random number $r \in [0,1)$}
        \If{$r < p_{mask}[i]$}
          \State $\tilde{T_f}[i][j] = [mask]$
          \Comment{Replace token with mask embedding based on probability}
        \EndIf
      \EndFor
    \EndFor

    \State $T_t = \text{Concat}(T_h, T_f)$
    \Comment{Concatenate historical and future tokens for supervision signal}

    \State $\tilde{T_t} = \text{Concat}(T_h, \tilde{T_f})$
    \Comment{Concatenate historical and partially masked future tokens for model input}

    \State $Z = \text{Embed}(\tilde{T_t}, \hat{E},E_{mask})$
    \Comment{Construct continuous input $Z$ using token indices and vocabularies}

    \State $\hat{Z} = \text{TRM}(Z)$
    \Comment{Pass the input signal $Z$ through the Backbone to get output $\hat{Z}$}

    \State $\mathcal{L} = \mathcal{L}_{diffusion}(\hat{Z}, T_t)$
    \Comment{Compute loss between the output $\hat{Z}$ and ground truth $T_t$}
    
    \State \Return $\mathcal{L}$
    \Comment{Return the loss for optimization}
  \end{algorithmic}
  \label{appendix_alg:train}
\end{algorithm*}

\begin{algorithm*}[htbp]
  \small
  \caption{Inference Process of Token Predictor}\label{algo:tokeninfer}
  \begin{algorithmic}[1]
    \Require  
   Historical tokens \( T_h \in \mathbb{N}^{B \times L_h}\), future token length \( L_f \), mask embedding $E_{mask} \in \mathbb{R}^{1 \times D}$, number of inference steps \( T \), and pretrained backbone TRM.

    \State $\tilde{T_f} = \text{CreateMatrixOfMaskID}(B, L_h)$
    \Comment{Initialize $\tilde{T_f} \in \mathbb{N}^{B \times L_h}$ with mask tokens}

    \State $\tilde{T_t} = \text{Concat}(T_h, \tilde{T_f})$
    \Comment{Set initial input by concatenating historical tokens and masked future tokens}

    \State $N = L_f // T$
    \Comment{Calculate number of tokens to restore in each round}

    \For{each inference step $k \in \{1, \ldots, T\}$}
      \State $Z = \text{Embed}(\tilde{T_t}, \hat{E}, E_{mask})$
      \Comment{Construct continuous input signal $Z$}

      \State $\hat{Z} = \text{TRM}(Z)$
      \Comment{Pass the input signal $Z$ through the Backbone to get output $\hat{Z}$}

      \State $\text{probabilities} = \text{Softmax}(\hat{Z})$
      \Comment{Compute probabilities for each token position}

      \State $\text{mask\_positions} = \text{FindIndices}(\tilde{T_f} == [mask])$
      \Comment{Identify positions still marked as mask in $\tilde{T_f}$}

      \State $\text{top\_positions} = \text{SelectTopN}(\text{probabilities}, \text{mask\_positions}, N)$
      \Comment{Select top $N$ positions with highest probabilities among mask positions}

      \State $\tilde{T_f}[\text{top\_positions}] = \text{RestoreTokens}(\text{probabilities}, \text{top\_positions})$
      \Comment{Restore tokens at selected positions}

      \State $\tilde{T_t} = \text{Concat}(T_h, \tilde{T_f})$
      \Comment{Update input for the next round}
    \EndFor

    \State $\hat{T_f} = \tilde{T_f}$
    \Comment{Predicted sequence of future tokens after $T$ rounds of denoising}

    \State \Return $\hat{T_f}$
    \Comment{Return the predicted future tokens}
  \end{algorithmic}
  \label{appendix_alg:infer}
\end{algorithm*}

%% file: appendix/appendix_C.tex
\section{Evidence Lower Bound of Discrete Diffusion}
\label{appendixC:elbo_discrete_diffusion}

Diffusion Models are a class of generative models that transform input data into latent variables by progressively injecting noise. This procedure is defined as a forward process, which is a Markov chain that begins with the original data \( \mathbf{x}_0 \) and generates a sequence of latent variables \( \mathbf{x}_1, \mathbf{x}_2, \ldots, \mathbf{x}_T \) with increasing levels of noise. The central objective of the model is to learn a corresponding reverse process, which also follows the Markov assumption, aimed at systematically removing the noise to recover the original data distribution from the latent variables.

In the continuous data domain, the transition distribution of the forward process, \( q(\mathbf{x}_t | \mathbf{x}_{t-1}) \), is typically modeled as a Gaussian distribution with mean \( \sqrt{1 - \beta_t} \mathbf{x}_{t-1} \) and variance \( \beta_t \mathbf{I} \). This definition implies that at each timestep, a specific intensity of Gaussian noise is added to the data.

For discrete data domains, however, such as when processing time-series data into one-hot encoded vectors over \( K+1 \) categories (including a mask state), the forward process is defined by a stochastic transition matrix \( \mathbf{Q}_t \in \mathbb{R}^{(K+1) \times (K+1)} \). Specifically, the transition distribution is defined as a Categorical distribution:
\begin{equation}
     q(\mathbf{x}_t | \mathbf{x}_{t-1}) = \text{Cat}(\mathbf{x}_t | \mathbf{x}_{t-1} \mathbf{Q}_t),
     \label{appendix_eq:forward_step}
\end{equation}
where the matrix element \( [\mathbf{Q}_t]_{ij} = q(\mathbf{x}_t = j | \mathbf{x}_{t-1} = i) \) represents the probability of transitioning from state \( i \) to state \( j \). Given the one-hot vector \( \mathbf{x}_{t-1} \), the product \( \mathbf{x}_{t-1} \mathbf{Q}_t \) yields a probability vector that defines the distribution of \( \mathbf{x}_t \).

Leveraging the properties of a Markov chain, the transition distribution from the initial state \( \mathbf{x}_0 \) directly to timestep \( t \) can be expressed concisely as:
\begin{equation}
      q(\mathbf{x}_t | \mathbf{x}_0) = \text{Cat}(\mathbf{x}_t | \mathbf{x}_0 \overline{\mathbf{Q}}_t),
      \label{appendix_eq:forward_marginal}
\end{equation}
where \( \overline{\mathbf{Q}}_t = \mathbf{Q}_1 \mathbf{Q}_2 \cdots \mathbf{Q}_t \) is the cumulative transition matrix. Based on Equations~\ref{appendix_eq:forward_step} and \ref{appendix_eq:forward_marginal}, we can derive the posterior distribution \( q(\mathbf{x}_{t-1} | \mathbf{x}_t, \mathbf{x}_0) \) using Bayes' theorem:
\begin{equation}
    \begin{aligned}
        q(\mathbf{x}_{t-1} | \mathbf{x}_t, \mathbf{x}_0) &= \frac{q(\mathbf{x}_t | \mathbf{x}_{t-1}, \mathbf{x}_0) q(\mathbf{x}_{t-1} | \mathbf{x}_0)}{q(\mathbf{x}_t | \mathbf{x}_0)} \\
        &= \text{Cat}\left(\mathbf{x}_{t-1} \middle| \frac{\mathbf{x}_t \mathbf{Q}_t^\top \odot \mathbf{x}_0 \overline{\mathbf{Q}}_{t-1}}{\mathbf{x}_0 \overline{\mathbf{Q}}_t \mathbf{x}_t^\top}\right), 
    \end{aligned}
\end{equation}
where \( \odot \) denotes the element-wise product, and the resulting vector parameterizes the categorical distribution for \( \mathbf{x}_{t-1} \).

The model is trained by maximizing the Evidence Lower Bound (ELBO) on the data log-likelihood. The ELBO is expressed as:
\begin{equation}
    \begin{aligned}
        \mathcal{L}_{\text{ELBO}}(\mathbf{x}_0, \theta) &= \mathbb{E}_{q(\mathbf{x}_{1:T}|\mathbf{x}_0)}\left[
        D_{\text{KL}}\big(q(\mathbf{x}_T|\mathbf{x}_0) \parallel p_\theta(\mathbf{x}_T)\big) - \log p_\theta(\mathbf{x}_0|\mathbf{x}_1) \right] \\
        &+ \mathbb{E}_{q(\mathbf{x}_{1:T}|\mathbf{x}_0)} \left[ \sum_{t=2}^{T} D_{\text{KL}}\big(q(\mathbf{x}_{t-1}|\mathbf{x}_t, \mathbf{x}_0) \parallel p_\theta(\mathbf{x}_{t-1}|\mathbf{x}_t)\big)
        \right],
    \end{aligned}
\end{equation}
where \( p_\theta \) is the reverse process parameterized by a neural network. As shown by~\cite{zhang2023copilot4d}, under the initial distribution \( q(\mathbf{x}_0) \), this variational lower bound can be further simplified into a reconstruction-oriented objective:
\begin{equation}
    \begin{aligned}
    \mathbb{E}_{q(\mathbf{x}_0)} \left[ \log p_\theta(\mathbf{x}_0) \right] & \geq \mathbb{E}_{q(\mathbf{x}_0)} \left[ -\mathcal{L}_{\text{ELBO}}(\mathbf{x}_0, \theta) \right] \\
    & \geq \sum_{t=1}^{T} \mathbb{E}_{q(\mathbf{x}_0, \mathbf{x}_t)} \left[ \log p_\theta(\mathbf{x}_0 | \mathbf{x}_t) \right] + C,
    \end{aligned}
\end{equation}
where C is a constant that does not depend on \( \theta \). This formulation reveals that maximizing the lower bound is equivalent to maximizing the expected log-likelihood of reconstructing the original data \( \mathbf{x}_0 \) from its noised version \( \mathbf{x}_t \) across all timesteps.

To further simplify this discrete diffusion framework for practical applications,~\cite{xie2024show} introduces a special absorbing state transition matrix \( \mathbf{Q}_t \):
\begin{equation}
    [\mathbf{Q}_t]_{i,j} = 
    \begin{cases} 
    1, & \text{if } i = j = \text{[M]}, \\
    \beta_t, & \text{if } j = \text{[M]}, i \neq \text{[M]}, \\
    1 - \beta_t, & \text{if } i = j \neq \text{[M]}, \\
    0, & \text{otherwise},
    \end{cases}
\end{equation}
where \(\text{[M]}\) represents the mask token, and \( \beta_t \) is the probability of converting a non-mask token into the mask token at timestep \( t \). The key advantage of this design is its simplification of state transitions: any token either remains unchanged or is replaced by the unique absorbing state \(\text{[M]}\), with no possibility of transitioning to other specific categories. This setup directly reframes the model's learning objective as a masked token prediction task. The goal of the neural network \( p_\theta(\mathbf{x}_0 | \mathbf{x}_t) \) thus becomes clear: to reconstruct the original \( \mathbf{x}_0 \) from the noised (i.e., partially masked) input \( \mathbf{x}_t \).

%% file: appendix/appendix_D.tex
\section{Details of Main Results}

\begin{table*}[h]
\centering
\tiny
\caption{Multivariate time series forecasting full results with historical window L=96, highlighting the best values in \textbf{bold} and the second-best values {\ul underlined}. The "$1^{\text{st}}$ Counts" indicates the times each method achieves the best results.}
\setlength{\tabcolsep}{2.5pt}
\resizebox{\textwidth}{!}{
\begin{tabular}{@{}cc|cccccccc|cccccccccccc@{}}
\toprule
\multicolumn{2}{c|}{Type} & \multicolumn{8}{c|}{Models Trained Across-Domain} & \multicolumn{12}{c}{Models Trained In-Domain} \\ 
\cmidrule(l){3-22} 
\multicolumn{2}{c|}{Methods} & \multicolumn{2}{c}{OneCast} & \multicolumn{2}{c}{UniTime} & 
\multicolumn{2}{c}{TOTEM} & \multicolumn{2}{c|}{TimesFM} &
\multicolumn{2}{c}{PatchTST} & \multicolumn{2}{c}{FEDformer} & \multicolumn{2}{c}{Autoformer} & 
\multicolumn{2}{c}{FILM} & \multicolumn{2}{c}{DLinear} &
\multicolumn{2}{c}{MICN} \\
\multicolumn{2}{c|}{Metric} & MSE & MAE & MSE & MAE & MSE & MAE & MSE & MAE & MSE & MAE & MSE & MAE & MSE & MAE & MSE & MAE & MSE & MAE & MSE & MAE \\
\midrule

\multicolumn{1}{c|}{\multirow{4}{*}{\rotatebox{90}{Traffic}}} & 24 & \textbf{0.469} & \underline{0.312} & 0.526 & 0.357 & 0.583 & 0.328 & 0.531 & 0.348 & 0.531 & 0.358 & 0.545 & 0.357 & 0.561 & 0.368 & 0.601 & 0.384 & 0.601 & \textbf{0.298} & \underline{0.492} & 0.372  \\
\multicolumn{1}{c|}{} & 48 & \textbf{0.496} & \underline{0.339} & 0.552 & 0.367 & 0.609 & 0.349 & 0.541 & 0.345 & 0.583 & 0.383 & 0.566 & 0.360 & 0.601 & 0.379 & 0.603 & 0.397 & 0.612 & \textbf{0.300} & \underline{0.497} & 0.377  \\
\multicolumn{1}{c|}{} & 96 & \textbf{0.498} & \underline{0.332} & 0.545 & 0.362 & 0.635 & 0.355 & 0.535 & 0.353 & 0.577 & 0.372 & 0.577 & 0.356 & 0.595 & 0.370 & 0.600 & 0.401 & 0.628 & \textbf{0.312} & \underline{0.520} & 0.362  \\
\multicolumn{1}{c|}{} & 192 & \textbf{0.505} & \underline{0.338} & 0.547 & 0.366 & 0.640 & 0.368 & 0.564 & 0.352 & 0.555 & 0.359 & 0.608 & 0.378 & 0.617 & 0.391 & 0.611 & 0.388 & 0.617 & \textbf{0.317} & \underline{0.535} & 0.368  \\ \midrule

\multicolumn{1}{c|}{\multirow{4}{*}{\rotatebox{90}{Weather}}} & 24 & \underline{0.113} & \textbf{0.151} & 0.125 & 0.166 & 0.130 & 0.183 & 0.129 & 0.158 & 0.119 & \underline{0.153} & 0.169 & 0.255 & 0.172 & 0.255 & 0.122 & 0.170 & 0.118 & 0.174 & \textbf{0.111} & 0.185 \\
\multicolumn{1}{c|}{} & 48 & \textbf{0.149} & \textbf{0.194} & 0.156 & 0.201 & 0.172 & 0.227 & 0.167 & 0.211 & \underline{0.151} & \underline{0.200} & 0.218 & 0.298 & 0.228 & 0.306 & 0.160 & 0.222 & 0.161 & 0.225 & 0.152 & 0.207  \\
\multicolumn{1}{c|}{} & 96 & \textbf{0.189} & \underline{0.236} & 0.196 & 0.238 & 0.197 & 0.239 & 0.205 & 0.236 & 0.194 & \textbf{0.234} & 0.240 & 0.320 & 0.253 & 0.325 & 0.196 & 0.253 & 0.195 & 0.249 & \underline{0.192} & 0.237  \\
\multicolumn{1}{c|}{} & 192 & 0.246 & 0.273 & \textbf{0.231} & \underline{0.265} & 0.246 & 0.280 & 0.253 & 0.273 & \textbf{0.231} & \textbf{0.263} & 0.282 & 0.342 & 0.294 & 0.350 & 0.239 & 0.298 & 0.238 & 0.289 & \underline{0.233} & 0.271 \\ \midrule

\multicolumn{1}{c|}{\multirow{4}{*}{\rotatebox{90}{CzeLan}}} & 24 & \textbf{0.149} & \textbf{0.219} & 0.184 & 0.255 & 0.165 & 0.248 & \underline{0.150} & 0.233 & 0.157 & \underline{0.224} & 0.177 & 0.268 & 0.394 & 0.424 & 0.266 & 0.315 & 0.213 & 0.299 & 0.185 & 0.335  \\
\multicolumn{1}{c|}{} & 48 & \textbf{0.201} & 0.274 & 0.226 & \underline{0.273} & 0.211 & 0.281 & 0.214 & 0.288 & \underline{0.209} & \textbf{0.270} & 0.256 & 0.336 & 0.646 & 0.548 & 0.307 & 0.384 & 0.291 & 0.341 & 0.234 & 0.360  \\
\multicolumn{1}{c|}{} & 96 & \textbf{0.221} & \textbf{0.265} & 0.256 & 0.292 & 0.249 & 0.299 & 0.242 & 0.292 & \underline{0.226} & \underline{0.275} & 0.250 & 0.324 & 0.526 & 0.493 & 0.269 & 0.359 & 0.288 & 0.376 & 0.289 & 0.313  \\
\multicolumn{1}{c|}{} & 192 & \textbf{0.253} & \textbf{0.285} & 0.290 & 0.328 & 0.286 & 0.319 & 0.278 & 0.302 & \underline{0.256} & \underline{0.294} & 0.304 & 0.363 & 0.733 & 0.598 & 0.306 & 0.425 & 0.391 & 0.432 & 0.394 & 0.341  \\ \midrule

\multicolumn{1}{c|}{\multirow{4}{*}{\rotatebox{90}{ETTh2}}} & 24 & \textbf{0.171} & \textbf{0.264} & 0.199 & 0.285 & 0.231 & 0.307 & 0.198 & 0.297 & \underline{0.174} & \textbf{0.264} & 0.227 & 0.324 & 0.273 & 0.359 & 0.206 & \underline{0.270} & 0.176 & 0.308 & 0.217 & 0.288  \\
\multicolumn{1}{c|}{} & 48 & \underline{0.237} & \textbf{0.313} & 0.244 & \underline{0.315} & 0.286 & 0.343 & \textbf{0.232} & 0.320 & 0.246 & 0.329 & 0.279 & 0.355 & 0.309 & 0.373 & 0.257 & 0.318 & 0.242 & 0.343 & 0.267 & 0.323  \\ 
\multicolumn{1}{c|}{} & 96 & 0.306 & 0.358 & \underline{0.298} & 0.352 & 0.352 & 0.383 & 0.315 & \underline{0.349} & \textbf{0.289} & \textbf{0.341} & 0.351 & 0.394 & 0.365 & 0.408 & 0.364 & 0.386 & 0.322 & 0.407 & 0.331 & 0.364  \\
\multicolumn{1}{c|}{} & 192 & 0.401 & 0.408 & 0.378 & \underline{0.400} & 0.423 & 0.429 & \underline{0.373} & \underline{0.400} & \textbf{0.366} & \textbf{0.391} & 0.433 & 0.442 & 0.443 & 0.451 & 0.413 & 0.475 & 0.476 & 0.489 & 0.507 & 0.420  \\ \midrule

\multicolumn{1}{c|}{\multirow{4}{*}{\rotatebox{90}{ETTm2}}} & 24 & \textbf{0.111} & \textbf{0.202} & 0.121 & 0.225 & 0.129 & 0.230 & 0.122 & 0.225 & \underline{0.113} & 0.210 & 0.129 & 0.239 & 0.148 & 0.257 & \textbf{0.111} & 0.210 & 0.119 & \underline{0.208} & 0.120 & 0.210  \\
\multicolumn{1}{c|}{} & 48 & 0.151 & 0.247 & 0.151 & 0.249 & 0.160 & 0.254 & 0.169 & 0.251 & \textbf{0.137} & \textbf{0.233} & 0.161 & 0.263 & 0.173 & 0.273 & 0.146 & 0.247 & 0.146 & \underline{0.240} & \underline{0.144} & 0.243  \\
\multicolumn{1}{c|}{} & 96 & 0.209 & 0.291 & 0.186 & 0.271 & 0.196 & 0.278 & 0.185 & 0.284 & \textbf{0.177} & \textbf{0.261} & 0.202 & 0.287 & 0.229 & 0.310 & \underline{0.184} & 0.292 & 0.193 & 0.282 & 0.185 & \underline{0.266}  \\
\multicolumn{1}{c|}{} & 192 & 0.268 & 0.318 & \underline{0.248} & 0.310 & 0.281 & 0.340 & 0.252 & 0.348 & \textbf{0.242} & \textbf{0.303} & 0.271 & 0.331 & 0.281 & 0.339 & \underline{0.248} & 0.361 & 0.283 & 0.360 & 0.287 & \underline{0.306}  \\ \midrule

\multicolumn{1}{c|}{\multirow{4}{*}{\rotatebox{90}{FRED-MD}}} & 24 & \textbf{26.32} & \textbf{0.872} & 44.26 & 1.182 & 46.65 & 1.304 & \underline{29.88} & \underline{0.888} & 39.61 & 1.100 & 66.58 & 1.636 & 66.72 & 1.714 & 63.37 & 1.621 & 69.69 & 2.044 & 70.87 & 1.478  \\
\multicolumn{1}{c|}{} & 36 & \textbf{48.69} & \textbf{0.940} & 71.45 & 1.484 & 57.98 & 1.482 & \underline{50.72} & \underline{1.111} & 68.23 & 1.447 & 94.91 & 1.883 & 95.09 & 1.953 & 106.1 & 1.905 & 99.74 & 1.965 & 106.3 & 1.882  \\
\multicolumn{1}{c|}{} & 48 & \underline{82.19} & \textbf{1.460} & 96.12 & 1.722 & \textbf{82.06} & 1.695 & 90.79 & \underline{1.543} & 100.1 & 1.747 & 130.4 & 2.138 & 130.6 & 2.195 & 130.3 & 2.220 & 137.4 & 2.652 & 171.8 & 2.087  \\
\multicolumn{1}{c|}{} & 60 & \underline{121.6} & \textbf{1.808} & 157.3 & 2.213 & \textbf{101.5} & \underline{1.834} & 132.2 & 1.851 & 142.7 & 2.112 & 174.4 & 2.436 & 174.6 & 2.489 & 175.4 & 2.531 & 182.0 & 3.107 & 231.0 & 2.413  \\ \midrule

\multicolumn{1}{c|}{\multirow{4}{*}{\rotatebox{90}{NYSE}}} & 24 & 0.269 & 0.369 & 0.245 & \underline{0.335} & 0.279 & 0.349 & \underline{0.234} & 0.342 & 0.305 & 0.362 & \textbf{0.226} & \textbf{0.330} & 0.427 & 0.466 & 0.439 & 0.573 & 0.600 & 0.584 & 0.496 & 0.448  \\
\multicolumn{1}{c|}{} & 36 & \textbf{0.324} & \textbf{0.353} & 0.399 & 0.428 & 0.442 & 0.445 & \underline{0.371} & \underline{0.410} & 0.411 & 0.418 & 0.414 & 0.447 & 0.526 & 0.508 & 0.566 & 0.702 & 0.856 & 0.573 & 0.640 & 0.512  \\
\multicolumn{1}{c|}{} & 48 & \textbf{0.481} & \textbf{0.427} & 0.533 & 0.491 & 0.573 & 0.505 & \underline{0.508} & \underline{0.459} & 0.639 & 0.532 & 0.631 & 0.550 & 0.757 & 0.605 & 0.716 & 0.802 & 1.112 & 0.796 & 1.064 & 0.583  \\
\multicolumn{1}{c|}{} & 60 & \textbf{0.654} & \textbf{0.517} & 0.776 & 0.613 & 0.753 & 0.597 & \underline{0.685} & \underline{0.535} & 0.820 & 0.614 & 0.800 & 0.626 & 0.994 & 0.714 & 1.015 & 0.938 & 1.447 & 0.988 & 1.513 & 0.720  \\ \midrule

\multicolumn{1}{c|}{\multirow{4}{*}{\rotatebox{90}{Covid-19}}} 
& 24 & \textbf{0.983} & \textbf{0.043} & 1.133 & 0.046 & 1.689 & 0.073 & \underline{1.086} & \underline{0.044} & 1.090 & 0.047 & 2.033 & 0.187 & 2.097 & 0.276 & 1.836 & 0.426 & 24.72 & 0.521 & 33.31 & 0.070 \\
\multicolumn{1}{c|}{} & 36 & \textbf{1.305} & \textbf{0.053} & 1.460 & \underline{0.054} & 2.249 & 0.093 & 1.450 & 0.055 & \underline{1.448} & 0.058 & 2.370 & 0.198 & 2.407 & 0.256 & 1.817 & 0.463 & 29.37 & 0.520 & 31.96 & 0.064 \\
\multicolumn{1}{c|}{} & 48 & \textbf{1.711} & 0.067 & 1.887 & \textbf{0.063} & 2.376 & 0.091 & \underline{1.775} & 0.069 & 1.812 & \underline{0.066} & 2.749 & 0.217 & 2.770 & 0.253 & 2.189 & 0.381 & 16.20 & 0.783 & 94.48 & 0.072 \\
\multicolumn{1}{c|}{} & 60 & \textbf{2.133} & \underline{0.075} & 2.359 & \textbf{0.073} & 2.966 & 0.093 & 2.267 & 0.078 & \underline{2.224} & \underline{0.075} & 3.165 & 0.233 & 3.193 & 0.277 & 2.687 & 0.551 & 44.06 & 0.927 & 130.6 & 0.083 \\ 
\midrule

\multicolumn{1}{c|}{\multirow{4}{*}{\rotatebox{90}{Wike2000}}} 
& 24 & \textbf{482.1} & \textbf{1.048} & 583.7 & 1.118 & 570.8 & 1.208 & \underline{506.8} & 1.163 & 524.5 & \underline{1.093} & 663.6 & 3.839 & 671.7 & 3.912 & 967.4 & 1.277 & 561.5 & 1.451 & 717.3 & 1.408 \\
\multicolumn{1}{c|}{} & 36 & \textbf{553.1} & \textbf{1.150} & 627.5 & \underline{1.217} & 813.4 & 1.341 & 578.4 & 1.379 & \underline{563.0} & 1.271 & 701.4 & 3.245 & 706.7 & 3.302 & 1171 & 1.362 & 630.1 & 1.363 & 616.9 & 1.533 \\
\multicolumn{1}{c|}{} & 48 & \textbf{580.9} & \textbf{1.223} & 649.6 & \underline{1.299} & 715.3 & 1.356 & \underline{590.7} & 1.347 & 604.3 & 1.314 & 734.5 & 3.092 & 736.9 & 3.076 & 1239 & 1.439 & 649.9 & 1.565 & 611.5 & 1.617 \\
\multicolumn{1}{c|}{} & 60 & \textbf{613.9} & \textbf{1.282} & 662.0 & 1.341 & \underline{616.4} & 1.360 & 653.3 & 1.353 & 647.6 & \underline{1.321} & 772.6 & 2.938 & 774.7 & 2.921 & 1337 & 1.514 & 687.7 & 1.706 & 637.3 & 1.672 \\ 

\midrule

\multicolumn{2}{c|}{$1^{\text{st}}$ Counts} & \multicolumn{2}{c}{46} & \multicolumn{2}{c}{3} & \multicolumn{2}{c}{2} & \multicolumn{2}{c}{1} & \multicolumn{2}{c}{15} & \multicolumn{2}{c}{2} & \multicolumn{2}{c}{0} & \multicolumn{2}{c}{1} & \multicolumn{2}{c}{4} & \multicolumn{2}{c}{1} \\
\bottomrule
\end{tabular}
}
\label{tab:full_table}
\end{table*}

We provide our complete forecasting results at Table \ref{tab:full_table}. As shown in table, OneCast achieves state-of-the-art results in 46 out of 72 entries when compared to other baselines, demonstrating competitive performance across both cross-domain and in-domain models.

%% file: appendix/appendix_E.tex
\section{More Analysis about Time Series Tokenizer}
\label{appendixE}


\subsection{Necessity of Dual-decoder in Training} 

In the process of encoding time series components, we apply instance normalization to time series segments to mitigate the issue of distribution shift \cite{kim2021reversible}. Specifically, for a segment of historical series $X_{h}$ and a segment of future series $X_{f}$, the normalization operation is as follows:

\begin{equation}
    X_{h,norm} = \frac{X_{h} - \mu_{h}}{\sqrt{\sigma^2_{h} + \epsilon}} ,     X_{f,norm} = \frac{X_{f} - \mu_{f}}{\sqrt{\sigma^2_{f} + \epsilon}},
\end{equation}

\noindent where $\epsilon$ is a small constant used to prevent division by zero errors, $\mu_{h}, \mu_{f}, \sigma_{h}$ and $ \sigma_{f} \in \mathbb{R}^{C}$ are the mean and variance of historical and future windows in different channels, respectively. Subsequently, we perform the moving average $MA$ on \(X_{h,norm}\) and \(X_{f,norm}\) to obtain trend components \(X_{h,trend}\) and \(X_{f,trend}\), as well as seasonal components \(X_{h,season}\) and \(X_{f,season}\):

\begin{equation}
    \begin{aligned}
        MA_t &= \frac{1}{n} \sum_{i=0}^{n-1} x_{t-i}, t \in \{1, 2, ..., T\},\\
    \end{aligned}
    \label{eq:moving_avg}
\end{equation}

\noindent where $X = \{x_{1}, x_2, ..., x_T\}$ is the input series to be decomposition, $n$ represents the size of the sliding window, and padding will be used when \( t -i \leq 0 \). Following this, the trend component $X_{trend}$ and seasonal component $X_{season}$ will be calculated by:
\begin{equation}
    \begin{aligned}
        X_{trend} &= \{MA_i|i = {1, 2, ..., T}\}, \\
        X_{season} &= X - X_{trend}.
    \end{aligned}
\end{equation}

To enable the tokenizer to effectively compress trend information into discrete tokens, we employ the loss function $\mathcal{L}_{1}$ mentioned in main text for optimization as follows:

\begin{equation}
    \mathcal{L}_{1} = \Vert\mathcal{D}_{(\mu_h,\sigma_h)}{X_{h,trend} - \mathcal{D}_{(\mu_h,\sigma_h)}\hat{X}_{h,trend}} \Vert^2 ,\\
    \label{appendix_eq:L1}
\end{equation}

\noindent where $\mathcal{D}_{(\mu,\sigma)}(\cdot)$ denotes denormalization using specific means $\mu$ and variances $\sigma$:

\begin{equation}
    \mathcal{D}_{(\mu, \sigma)}(X) \triangleq X \cdot \sigma + \mu .
\end{equation}

When using a single decoder, reconstructing future tokens requires the mean $\mu_f$ and variance $\sigma_f$ of the future window, which are unavailable during prediction and may differ markedly from historical statistics $(\mu_h, \sigma_h)$. To address this, we design a dual-decoder strategy. The historical decoder $\mathbf{D_h}$ reconstructs the historical trend window via loss $\mathcal{L}_1$ (Eq.~\ref{appendix_eq:L1}), encouraging the encoder and codebook to learn compact representations. The future decoder $\mathbf{D_f}$ leverages historical statistics to decode future tokens, mitigating distributional shift under a data-driven manner. Its optimization objective, $\mathcal{L}_{2}$, is as follows:


\begin{equation}
    \mathcal{L}_{2} = \Vert\mathcal{D}_{(\mu_f,\sigma_f)}X_{f,trend} - \mathcal{D}_{(\mu_hf,\sigma_h)}\hat{X}_{f,trend}\Vert^2.
\end{equation}

It is worth noting that to prevent $\mathcal{L}_{2}$ from interfering with the learning of effective token compression , the gradients introduced by $\mathcal{L}_{2}$ are truncated before being backpropagated to the codebook. 

\subsection{Feasibility of Decomposition Strategies}
\begin{figure}[ht]
    \centering
    \includegraphics[width=0.95\linewidth]{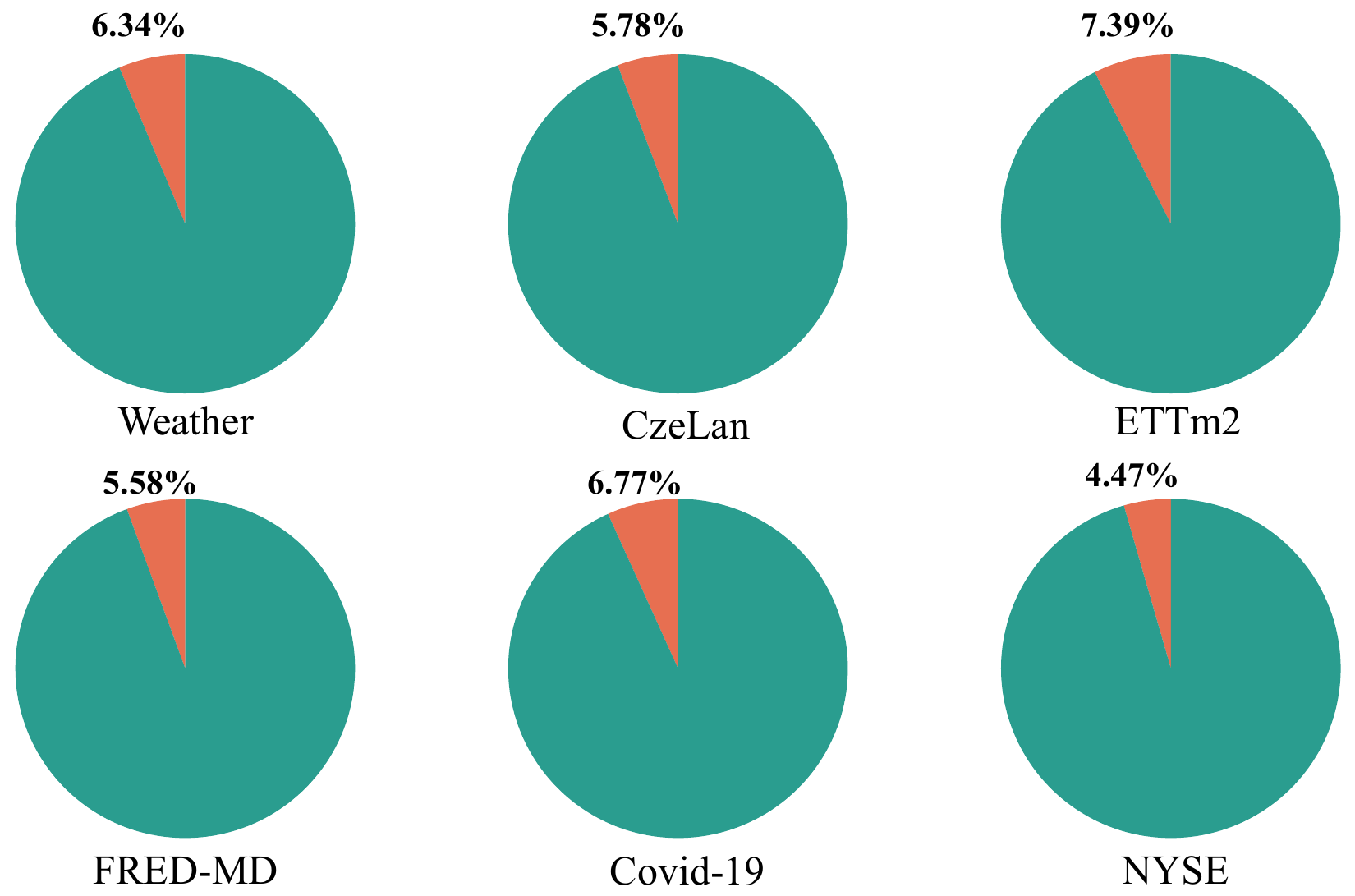}
    \caption{Residual component rate in different datasets.}
    \label{appendix_fig:res_comp}
\end{figure}

Time series are commonly decomposed into trend $X_{T}$, seasonality $X_{S}$, and residuals $X_{R}$:

\begin{equation}
X = X_{T} + X_{S} + X_{R},
\end{equation}

\noindent where residuals $X_{R}$ are generally treated as random noise and considered unpredictable. Although Moving Average provides an efficient decomposition, it cannot fully isolate residuals from seasonal variations, which may affect token modeling. To quantify the residual proportion, we define the onent rate (RCR):

\begin{equation}
\text{RCR}(X_{T}, X_{S}, X_{R}) = \frac{1}{N}\sum_{i=1}^N \frac{|x_{r,i}|}{|x_{t,i}|+|x_{s,i}|+|x_{r,i}|},
\end{equation}

\noindent where \( N \) denotes the number of data points, while \( x_{t,i} \), \( x_{s,i} \), and \( x_{r,i} \) represent the values of the trend, seasonal, and onent corresponding to the \( i \)-th data point, respectively. Using STL decomposition, we estimate components and compute RCR across datasets. Table \ref{appendix_fig:res_comp} shows that residuals account for less than 8\% of the total variance, indicating that random perturbations are minor, and trend and seasonality dominate the series. This supports the adequacy of Moving Average for decomposition.







\subsection{Performance Balance Between Tokenizer and Token Predictor}
\label{sec:balance_tokenizer_predictor}

We study how two granularity choices shape both encoding and downstream prediction: the \emph{patch length} $P$ (with stride $=$ patch, i.e., non-overlapping segmentation) and the \emph{wave length} $W$ (the token span, i.e., the number of time steps covered by one token).
Given a window of length $L$, the number of patches is $L/P$, each patch yields $P/W$ tokens, and thus the \emph{total number of tokens} satisfies:

\begin{equation}
N_{\text{tok}} \;=\; \frac{L}{P}\cdot\frac{P}{W} \;=\; \frac{L}{W}\\,
\end{equation}

\noindent which depends only on $W$ under the non-overlapping assumption and divisibility.


\subsubsection{Varying Patch Length $P$.}
\begin{figure}[ht]
    \centering
    \includegraphics[width=\linewidth]{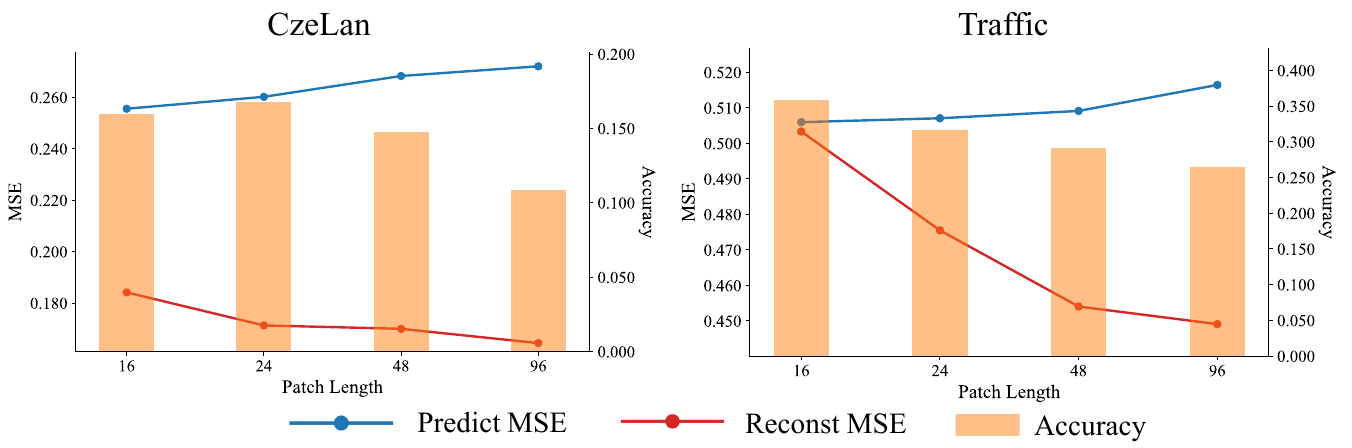}
    \caption{Effect of patch length $P\in\{16,24,48,96\}$ with fixed future length $L_f{=}192$ and wave length $W{=}8$.
    Orange bars: token predict accuracy; blue and red lines: forecast error and tokenizer reconstruction error on future window, respectively.}
    \label{appendix_fig:diff_tl}
\end{figure}

We fix $W{=}8$ and compare $P\in\{16,24,48,96\}$. In this setting, sequences of the same length are encoded into an equal number of tokens ($N_{\text{tok}}{=}L_f/W$), while the \emph{tokens per patch} change as $\{2,3,6,12\}$.
As shown in Figure \ref{appendix_fig:diff_tl}, increasing $P$ (i.e., encoding larger units) basically \emph{reduces} the tokenizer’s reconstruction error on the future window. Intuitively, longer units provide richer intra-patch context to the tokenizer, facilitating reconstruction. However, despite $N_{\text{tok}}$ being constant, the \emph{token prediction accuracy} of the downstream classifier \emph{decreases} as $P$ grows: each token aggregates more heterogeneous temporal content, making the generation task intrinsically harder. This directly challenges the intuition that a stronger tokenizer (in reconstruction) necessarily yields better final forecasting.

\subsubsection{Varying Wave Length $W$.}
We fix $P{=}96$ and compare $W\in\{4,8,16,32\}$, which changes tokens per patch to $\{24,12,6,3\}$.
Figure \ref{appendix_fig:diff_wl} shows a characteristic trade-off: the tokenizer’s reconstruction error often exhibits a U-shaped trend w.r.t.\ $W$—shorter $W$ encodes more tokens with higher redundancy and finer granularity, while larger $W$ benefits from wider receptive fields until overly coarse granularity starts to lose fine details. In contrast, the downstream \emph{token prediction accuracy} improves steadily with larger $W$ because fewer tokens render an easier classification problem.

\subsubsection{Implication.}
\begin{figure}[ht]
    \centering
    \includegraphics[width=\linewidth]{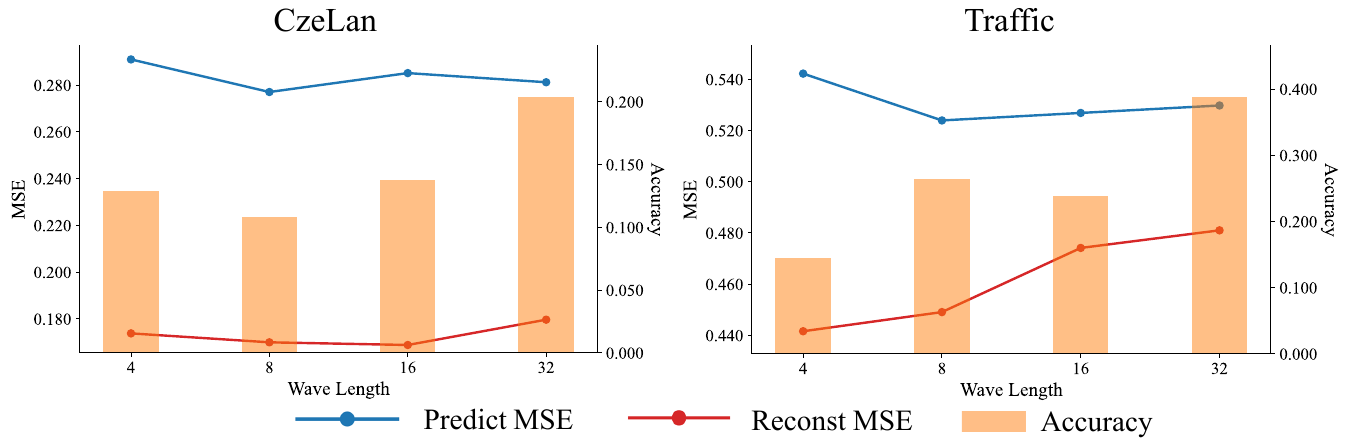}
    \caption{Effect of wave length $W\in\{4,8,16,32\}$ with fixed future length $L_f{=}192$ and patch length $P{=}96$.
    Orange bars, blue and red lines are the same meaning as mentioned above.}
    \label{appendix_fig:diff_wl}
\end{figure}

Across Figure \ref{appendix_fig:diff_tl} and Figure \ref{appendix_fig:diff_wl}, the \emph{final} forecasting error on the future window does not follow a simple monotone relation with either tokenizer reconstruction or token prediction accuracy. In other words, high-fidelity reconstruction at the tokenization stage does not guarantee superior downstream forecasting. Practically, it should be \emph{balanced} between the tokenizer’s encoding and decoding capacity and the learnability of the token predictor. Tuning $(P,W)$ to control token granularity and $N_{\text{tok}}$ is essential to align representational fidelity with predictive difficulty.


%% file: appendix/appendix_F.tex
\section{Ablation Study}


\subsection{Varying Codebook Size of Tokenizer}

\begin{figure}[ht]
    \centering
    \includegraphics[width=\linewidth]{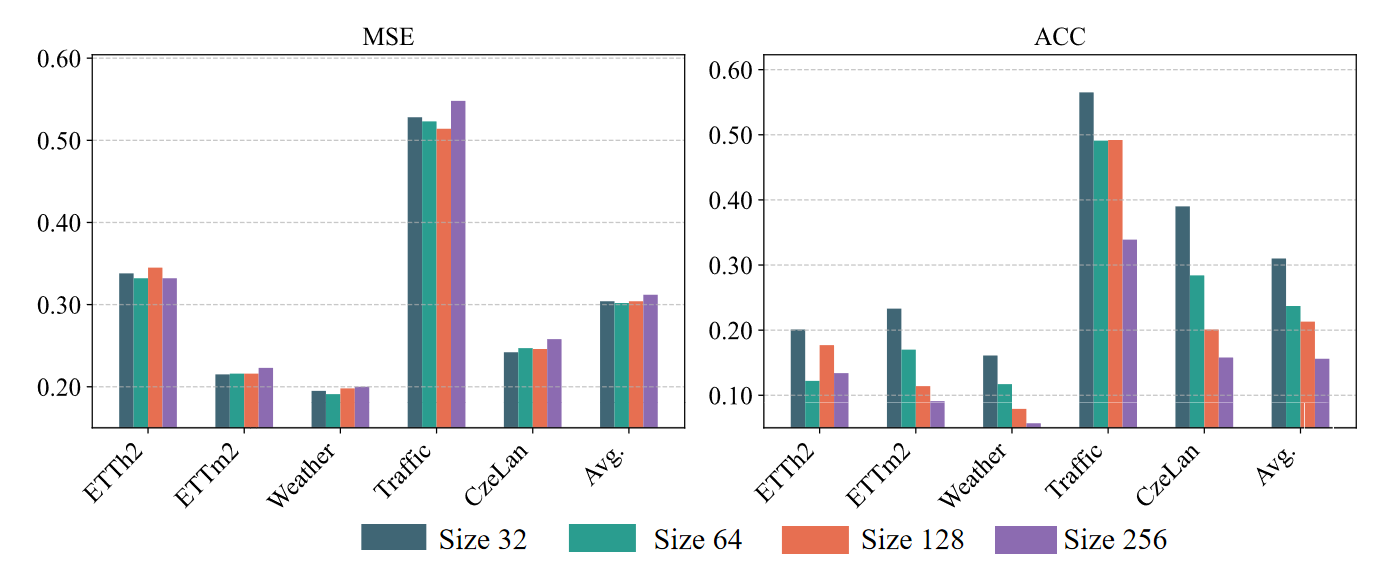}
    \caption{Comparison of codebook sizes with future length 96. MSE measures prediction errors, while ACC denotes the token predictor’s accuracy on future sequence.}
    \label{appendix_fig:diff_codesize}
\end{figure}

To investigate the impact of different codebook sizes on experimental results, we conduct comparative experiments on five large datasets, selecting a prediction window length of 96 as a central value. Figure \ref{appendix_fig:diff_codesize} presents the corresponding experimental results. It is evident that varying codebook sizes significantly affect both MSE and MAE, particularly in the Traffic dataset. Furthermore, as the codebook size increases, the prediction accuracy for temporal tokens declines sharply. This decline may be attributed to the increased complexity faced by downstream predictors as the vocabulary size expands.

\subsection{Varying Token Abandonment Rates}
As shown in Figure \ref{appendix_fig:diff_codesize}, different codebook sizes strongly affect token prediction accuracy, but have limited impact on MSE and MAE. To examine this, we fix the codebook size at 128 and discard tokens with training frequencies below 0\%, 0.1\%, 0.4\%, 0.7\%, and 1\%. Figure \ref{appendix_fig:diff_freq} shows that higher abandonment thresholds consistently improve token accuracy, while prediction errors remain stable on most datasets but fluctuate on Traffic. This may result from its larger channel dimension, which demands more effective tokens for compression—reducing tokens weakens reconstruction yet benefits downstream prediction, leading to fluctuating errors.

\subsection{Varying Inference Step of Diffusion}

To thoroughly demonstrate the impact of the number of inference steps, we set the future window length to 192 and experiment with five different inference steps: 1, 4, 8, 12, and 16. As shown in Table \ref{appendix_tab:infer_step}, the predictions obtained after 4 inference steps exhibit a significant advantage. Although 1-step inference performs well on the Traffic and CzeLan datasets, it underperforms on the other three datasets, resulting in an unsatisfactory overall average. This indicates that 1-step inference may be less stable compared to multi-step inference, and excessive inference steps can directly lead to a decline in performance.

\subsection{Varying Historical Length}

Theoretically, longer histories provide richer context, but attention models can suffer dilution on very long sequences \cite{wang2024timexer}. We therefore evaluate OneCast with varying input lengths across four horizons. As shown in Figure \ref{appendix_fig:diff_input}), OneCast attains its ideal performance on Traffic, whereas an input length of 336 is optimal for the other datasets. Extending the history to 720 consistently degrades accuracy, suggesting OneCast lacks the capability to effectively understand long sequence, motivating future work on long-context modeling.

%% file: appendix/appendix_G.tex
\section{Visualization}

\begin{table*}[h]
    \centering
    \large
    \caption{Comparison of different inference steps, under the future window length of 192.}
    \vspace{-0.1in}
    \begin{tabular}{@{}c|cc|cc|cc|cc|cc|c@{}}
        \toprule
         \multirow{2}{*}{Steps} & \multicolumn{2}{c|}{ETTh2} & \multicolumn{2}{c|}{ETTm2} & \multicolumn{2}{c|}{Weather} & \multicolumn{2}{c|}{Traffic} &
        \multicolumn{2}{c|}{CzeLan} & \multirow{2}{*}{Avg.}\\
        \cmidrule(lr){2-11}
         & mse & mae & mse & mae & mse & mae & mse & mae & mse & mae &\\
        \midrule
        1 & 0.442 & 0.433 & 0.290 & 0.333 & 0.267 & 0.292 & \textbf{0.528} & \textbf{ 0.356} & \textbf{0.273} & \textbf{0.306} & 0.352\\
        4 & \textbf{0.438} & \textbf{0.426} & \textbf{0.278} & \textbf{0.328} & \textbf{0.262} & \textbf{0.284} & 0.530 & 0.357 & 0.277 & 0.307 & \textbf{0.349}\\
        8 & 0.445 & 0.430 & 0.286 & 0.332 & 0.264 & \textbf{0.284} & 0.530 & 0.357 & 0.277 & 0.307 & 0.351\\
        12 & 0.449 & 0.431 & 0.291 & 0.335 & 0.264 & 0.285 & 0.530 & 0.358 & 0.277 & 0.308 & 0.353\\
        16 & 0.457 & 0.435 & 0.295 & 0.338 & 0.264 & 0.285 & 0.530 & 0.358 & 0.278 & 0.308 & 0.355\\
        \bottomrule
    \end{tabular}
    \label{appendix_tab:infer_step}
\end{table*}

\begin{figure*}[ht]
    \centering
    \includegraphics[width=\textwidth]{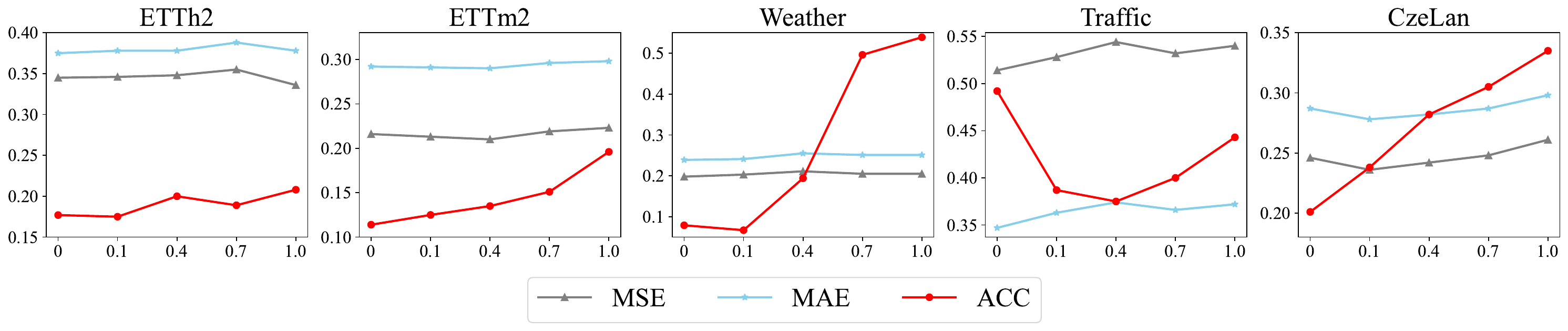}
    \caption{Comparison of different token abandonment rates, under the future window length of 96. MSE, MAE and ACC are the same meanings as Figure \ref{appendix_fig:diff_codesize}.}
    \label{appendix_fig:diff_freq}
\end{figure*}

\begin{figure*}[ht]
    \centering
    \includegraphics[width=\textwidth]{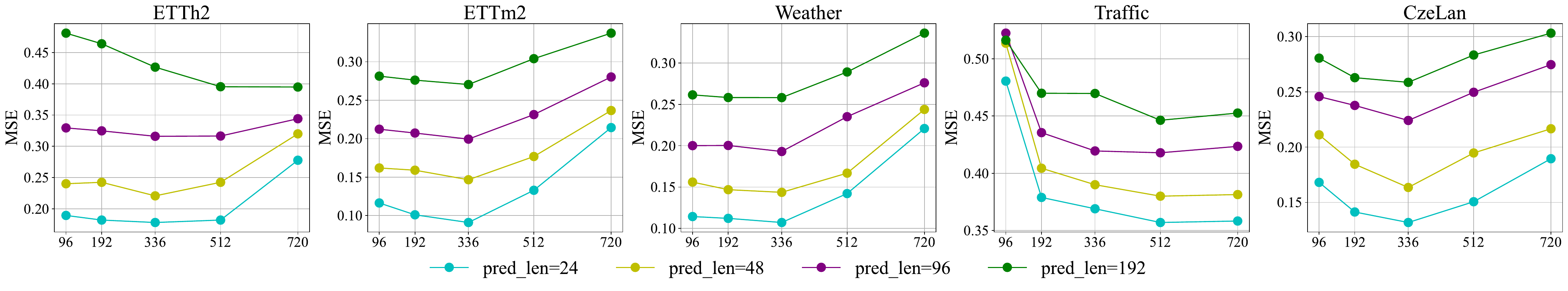}
    \caption{Performance with increasing historical window length, ranging from \{96, 192, 336, 512, 720\}. Different line styles denote different prediction lengths.}
    \label{appendix_fig:diff_input}
\end{figure*}

\subsection{Visualization of Prediction Results}
\begin{figure*}[ht]
    \centering
    \includegraphics[width=\textwidth]{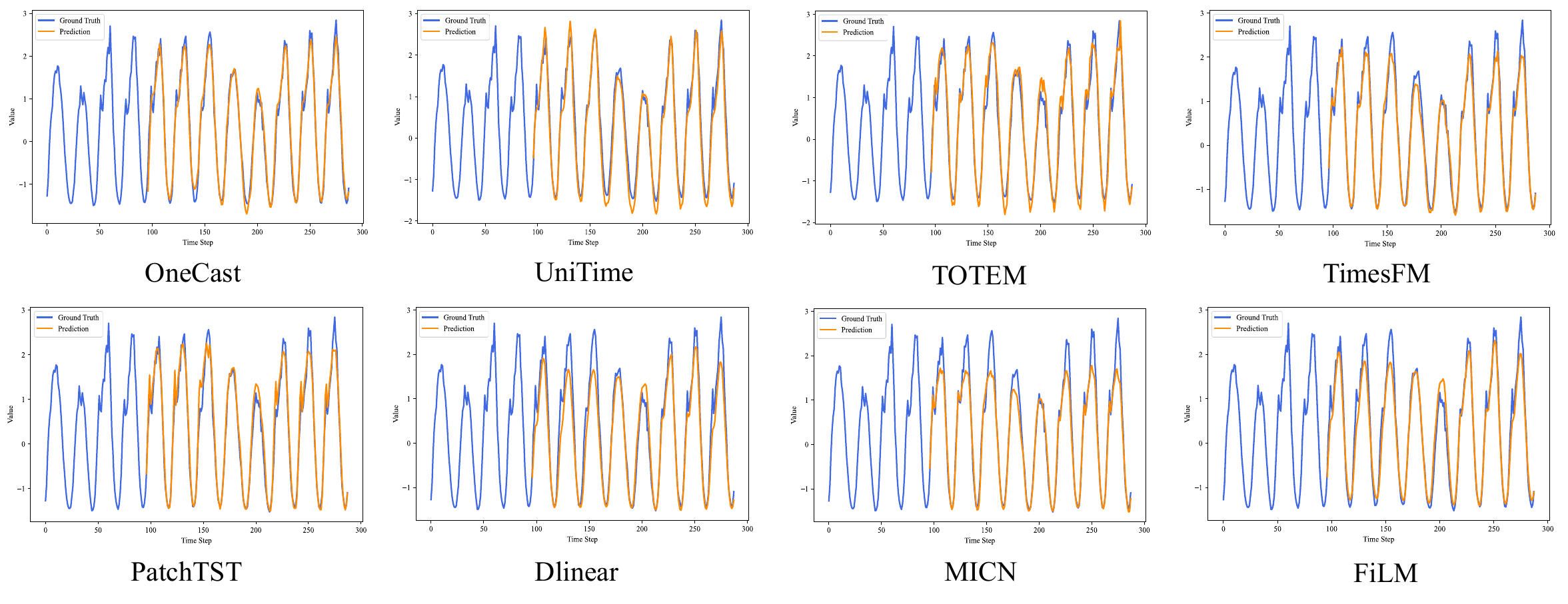}
    \caption{Forecasting showcase of OneCast and baseline models, where the blue line represents the ground truth, while the orange line shows the predicted values produced by the corresponding model.}
    \label{appendix_fig:case}
\end{figure*}

In the visualization shown on Figure \ref{appendix_fig:case},  we demonstrate the advantages of OneCast compared to other baselines on a traffic case with historical window length of 96 and future window length of 192. In comparison, OneCast not only accurately captures the changes in peak values but also precisely predicts the magnitude of each peak, providing smooth predictions that are closest to the ground truth.

\subsection{Visualization of Reconstruction Results}
\begin{figure*}[htbp]
    \centering
    \includegraphics[width=\textwidth]{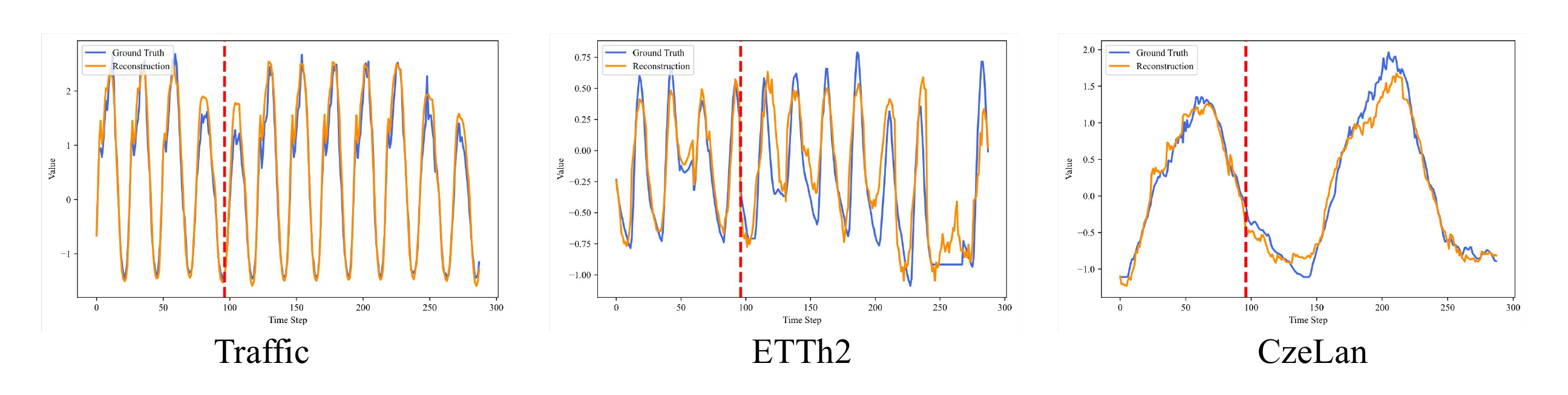}
    \caption{Reconstruction showcase of the time series tokenizer, where the blue line represents the ground truth, and the orange line illustrates the reconstructed values generated by the Tokenizer. The red dashed line indicates the boundary between the historical window and the future window.}
    \label{appendix_fig:reconst}
\end{figure*}

Figure \ref{appendix_fig:reconst} illustrates the compression capability of Our Tokenizer for time series data and its decoding ability for future windows. It is evident that, for the historical window on the left side of the red dashed line, the tokenizer can reconstruct the sequence with minimal information loss, given that the statistical properties are known. Conversely, for the future window on the right side of the dashed line, it effectively decodes the sequence even in the absence of knowledge about the statistical properties.

%% file: main.bbl

\begin{thebibliography}{49}


\ifx \showCODEN    \undefined \def \showCODEN     #1{\unskip}     \fi
\ifx \showISBNx    \undefined \def \showISBNx     #1{\unskip}     \fi
\ifx \showISBNxiii \undefined \def \showISBNxiii  #1{\unskip}     \fi
\ifx \showISSN     \undefined \def \showISSN      #1{\unskip}     \fi
\ifx \showLCCN     \undefined \def \showLCCN      #1{\unskip}     \fi
\ifx \shownote     \undefined \def \shownote      #1{#1}          \fi
\ifx \showarticletitle \undefined \def \showarticletitle #1{#1}   \fi
\ifx \showURL      \undefined \def \showURL       {\relax}        \fi
\providecommand\bibfield[2]{#2}
\providecommand\bibinfo[2]{#2}
\providecommand\natexlab[1]{#1}
\providecommand\showeprint[2][]{arXiv:#2}

\bibitem[Ansari et~al\mbox{.}(2024)]%
        {ansari2024chronos}
\bibfield{author}{\bibinfo{person}{Abdul~Fatir Ansari}, \bibinfo{person}{Lorenzo Stella}, \bibinfo{person}{Caner Turkmen}, \bibinfo{person}{Xiyuan Zhang}, \bibinfo{person}{Pedro Mercado}, \bibinfo{person}{Huibin Shen}, \bibinfo{person}{Oleksandr Shchur}, \bibinfo{person}{Syama~Sundar Rangapuram}, \bibinfo{person}{Sebastian~Pineda Arango}, \bibinfo{person}{Shubham Kapoor}, {et~al\mbox{.}}} \bibinfo{year}{2024}\natexlab{}.
\newblock \showarticletitle{Chronos: Learning the language of time series}.
\newblock \bibinfo{journal}{\emph{arXiv preprint arXiv:2403.07815}} (\bibinfo{year}{2024}).
\newblock


\bibitem[Casado-Vara et~al\mbox{.}(2021)]%
        {casado2021web}
\bibfield{author}{\bibinfo{person}{Roberto Casado-Vara}, \bibinfo{person}{Angel Martin~del Rey}, \bibinfo{person}{Daniel P{\'e}rez-Palau}, \bibinfo{person}{Luis de-la Fuente-Valent{\'\i}n}, {and} \bibinfo{person}{Juan~M Corchado}.} \bibinfo{year}{2021}\natexlab{}.
\newblock \showarticletitle{Web traffic time series forecasting using LSTM neural networks with distributed asynchronous training}.
\newblock \bibinfo{journal}{\emph{Mathematics}} \bibinfo{volume}{9}, \bibinfo{number}{4} (\bibinfo{year}{2021}), \bibinfo{pages}{421}.
\newblock


\bibitem[Cheng et~al\mbox{.}(2025a)]%
        {cheng2025comprehensive}
\bibfield{author}{\bibinfo{person}{Mingyue Cheng}, \bibinfo{person}{Zhiding Liu}, \bibinfo{person}{Xiaoyu Tao}, \bibinfo{person}{Qi Liu}, \bibinfo{person}{Jintao Zhang}, \bibinfo{person}{Tingyue Pan}, \bibinfo{person}{Shilong Zhang}, \bibinfo{person}{Panjing He}, \bibinfo{person}{Xiaohan Zhang}, \bibinfo{person}{Daoyu Wang}, {et~al\mbox{.}}} \bibinfo{year}{2025}\natexlab{a}.
\newblock \showarticletitle{A comprehensive survey of time series forecasting: Concepts, challenges, and future directions}.
\newblock \bibinfo{journal}{\emph{Authorea Preprints}} (\bibinfo{year}{2025}).
\newblock


\bibitem[Cheng et~al\mbox{.}(2025b)]%
        {cheng2025cross}
\bibfield{author}{\bibinfo{person}{Mingyue Cheng}, \bibinfo{person}{Xiaoyu Tao}, \bibinfo{person}{Qi Liu}, \bibinfo{person}{Hao Zhang}, \bibinfo{person}{Yiheng Chen}, {and} \bibinfo{person}{Defu Lian}.} \bibinfo{year}{2025}\natexlab{b}.
\newblock \showarticletitle{Cross-Domain Pre-training with Language Models for Transferable Time Series Representations}. In \bibinfo{booktitle}{\emph{Proceedings of the Eighteenth ACM International Conference on Web Search and Data Mining}}. \bibinfo{pages}{175--183}.
\newblock


\bibitem[Cheng et~al\mbox{.}(2024)]%
        {cheng2024convtimenet}
\bibfield{author}{\bibinfo{person}{Mingyue Cheng}, \bibinfo{person}{Jiqian Yang}, \bibinfo{person}{Tingyue Pan}, \bibinfo{person}{Qi Liu}, {and} \bibinfo{person}{Zhi Li}.} \bibinfo{year}{2024}\natexlab{}.
\newblock \showarticletitle{Convtimenet: A deep hierarchical fully convolutional model for multivariate time series analysis}.
\newblock \bibinfo{journal}{\emph{arXiv preprint arXiv:2403.01493}} (\bibinfo{year}{2024}).
\newblock


\bibitem[Darlow et~al\mbox{.}(2024)]%
        {darlow2024dam}
\bibfield{author}{\bibinfo{person}{Luke Darlow}, \bibinfo{person}{Qiwen Deng}, \bibinfo{person}{Ahmed Hassan}, \bibinfo{person}{Martin Asenov}, \bibinfo{person}{Rajkarn Singh}, \bibinfo{person}{Artjom Joosen}, \bibinfo{person}{Adam Barker}, {and} \bibinfo{person}{Amos Storkey}.} \bibinfo{year}{2024}\natexlab{}.
\newblock \showarticletitle{Dam: Towards a foundation model for time series forecasting}.
\newblock \bibinfo{journal}{\emph{arXiv preprint arXiv:2407.17880}} (\bibinfo{year}{2024}).
\newblock


\bibitem[Das et~al\mbox{.}(2024)]%
        {das2024decoder}
\bibfield{author}{\bibinfo{person}{Abhimanyu Das}, \bibinfo{person}{Weihao Kong}, \bibinfo{person}{Rajat Sen}, {and} \bibinfo{person}{Yichen Zhou}.} \bibinfo{year}{2024}\natexlab{}.
\newblock \showarticletitle{A decoder-only foundation model for time-series forecasting}. In \bibinfo{booktitle}{\emph{Forty-first International Conference on Machine Learning}}.
\newblock


\bibitem[Ekambaram et~al\mbox{.}(2024)]%
        {ekambaram2024tiny}
\bibfield{author}{\bibinfo{person}{Vijay Ekambaram}, \bibinfo{person}{Arindam Jati}, \bibinfo{person}{Pankaj Dayama}, \bibinfo{person}{Sumanta Mukherjee}, \bibinfo{person}{Nam Nguyen}, \bibinfo{person}{Wesley~M Gifford}, \bibinfo{person}{Chandra Reddy}, {and} \bibinfo{person}{Jayant Kalagnanam}.} \bibinfo{year}{2024}\natexlab{}.
\newblock \showarticletitle{Tiny time mixers (ttms): Fast pre-trained models for enhanced zero/few-shot forecasting of multivariate time series}.
\newblock \bibinfo{journal}{\emph{Advances in Neural Information Processing Systems}}  \bibinfo{volume}{37} (\bibinfo{year}{2024}), \bibinfo{pages}{74147--74181}.
\newblock


\bibitem[Feng et~al\mbox{.}(2019)]%
        {feng2019temporal}
\bibfield{author}{\bibinfo{person}{Fuli Feng}, \bibinfo{person}{Xiangnan He}, \bibinfo{person}{Xiang Wang}, \bibinfo{person}{Cheng Luo}, \bibinfo{person}{Yiqun Liu}, {and} \bibinfo{person}{Tat-Seng Chua}.} \bibinfo{year}{2019}\natexlab{}.
\newblock \showarticletitle{Temporal relational ranking for stock prediction}.
\newblock \bibinfo{journal}{\emph{ACM Transactions on Information Systems (TOIS)}} \bibinfo{volume}{37}, \bibinfo{number}{2} (\bibinfo{year}{2019}), \bibinfo{pages}{1--30}.
\newblock


\bibitem[Gasthaus et~al\mbox{.}(2019)]%
        {gasthaus2019probabilistic}
\bibfield{author}{\bibinfo{person}{Jan Gasthaus}, \bibinfo{person}{Konstantinos Benidis}, \bibinfo{person}{Yuyang Wang}, \bibinfo{person}{Syama~Sundar Rangapuram}, \bibinfo{person}{David Salinas}, \bibinfo{person}{Valentin Flunkert}, {and} \bibinfo{person}{Tim Januschowski}.} \bibinfo{year}{2019}\natexlab{}.
\newblock \showarticletitle{Probabilistic forecasting with spline quantile function RNNs}. In \bibinfo{booktitle}{\emph{The 22nd international conference on artificial intelligence and statistics}}. PMLR, \bibinfo{pages}{1901--1910}.
\newblock


\bibitem[He et~al\mbox{.}(2022)]%
        {he2022diffusionbert}
\bibfield{author}{\bibinfo{person}{Zhengfu He}, \bibinfo{person}{Tianxiang Sun}, \bibinfo{person}{Kuanning Wang}, \bibinfo{person}{Xuanjing Huang}, {and} \bibinfo{person}{Xipeng Qiu}.} \bibinfo{year}{2022}\natexlab{}.
\newblock \showarticletitle{Diffusionbert: Improving generative masked language models with diffusion models}.
\newblock \bibinfo{journal}{\emph{arXiv preprint arXiv:2211.15029}} (\bibinfo{year}{2022}).
\newblock


\bibitem[Ho et~al\mbox{.}(2020)]%
        {ho2020denoising}
\bibfield{author}{\bibinfo{person}{Jonathan Ho}, \bibinfo{person}{Ajay Jain}, {and} \bibinfo{person}{Pieter Abbeel}.} \bibinfo{year}{2020}\natexlab{}.
\newblock \showarticletitle{Denoising diffusion probabilistic models}.
\newblock \bibinfo{journal}{\emph{Advances in neural information processing systems}}  \bibinfo{volume}{33} (\bibinfo{year}{2020}), \bibinfo{pages}{6840--6851}.
\newblock


\bibitem[Hu et~al\mbox{.}(2015)]%
        {hu2015web}
\bibfield{author}{\bibinfo{person}{Yan Hu}, \bibinfo{person}{Qimin Peng}, \bibinfo{person}{Xiaohui Hu}, {and} \bibinfo{person}{Rong Yang}.} \bibinfo{year}{2015}\natexlab{}.
\newblock \showarticletitle{Web service recommendation based on time series forecasting and collaborative filtering}. In \bibinfo{booktitle}{\emph{2015 ieee international conference on web services}}. IEEE, \bibinfo{pages}{233--240}.
\newblock


\bibitem[Hyndman et~al\mbox{.}(2008)]%
        {hyndman2008forecasting}
\bibfield{author}{\bibinfo{person}{Rob Hyndman}, \bibinfo{person}{Anne~B Koehler}, \bibinfo{person}{J~Keith Ord}, {and} \bibinfo{person}{Ralph~D Snyder}.} \bibinfo{year}{2008}\natexlab{}.
\newblock \bibinfo{booktitle}{\emph{Forecasting with exponential smoothing: the state space approach}}.
\newblock \bibinfo{publisher}{Springer Science \& Business Media}.
\newblock


\bibitem[Ke et~al\mbox{.}(2017)]%
        {ke2017lightgbm}
\bibfield{author}{\bibinfo{person}{Guolin Ke}, \bibinfo{person}{Qi Meng}, \bibinfo{person}{Thomas Finley}, \bibinfo{person}{Taifeng Wang}, \bibinfo{person}{Wei Chen}, \bibinfo{person}{Weidong Ma}, \bibinfo{person}{Qiwei Ye}, {and} \bibinfo{person}{Tie-Yan Liu}.} \bibinfo{year}{2017}\natexlab{}.
\newblock \showarticletitle{Lightgbm: A highly efficient gradient boosting decision tree}.
\newblock \bibinfo{journal}{\emph{Advances in neural information processing systems}}  \bibinfo{volume}{30} (\bibinfo{year}{2017}).
\newblock


\bibitem[Kim et~al\mbox{.}(2021)]%
        {kim2021reversible}
\bibfield{author}{\bibinfo{person}{Taesung Kim}, \bibinfo{person}{Jinhee Kim}, \bibinfo{person}{Yunwon Tae}, \bibinfo{person}{Cheonbok Park}, \bibinfo{person}{Jang-Ho Choi}, {and} \bibinfo{person}{Jaegul Choo}.} \bibinfo{year}{2021}\natexlab{}.
\newblock \showarticletitle{Reversible instance normalization for accurate time-series forecasting against distribution shift}. In \bibinfo{booktitle}{\emph{International conference on learning representations}}.
\newblock


\bibitem[Liu et~al\mbox{.}(2024a)]%
        {liu2024unitime}
\bibfield{author}{\bibinfo{person}{Xu Liu}, \bibinfo{person}{Junfeng Hu}, \bibinfo{person}{Yuan Li}, \bibinfo{person}{Shizhe Diao}, \bibinfo{person}{Yuxuan Liang}, \bibinfo{person}{Bryan Hooi}, {and} \bibinfo{person}{Roger Zimmermann}.} \bibinfo{year}{2024}\natexlab{a}.
\newblock \showarticletitle{Unitime: A language-empowered unified model for cross-domain time series forecasting}. In \bibinfo{booktitle}{\emph{Proceedings of the ACM Web Conference 2024}}. \bibinfo{pages}{4095--4106}.
\newblock


\bibitem[Liu et~al\mbox{.}(2024b)]%
        {liu2024moirai}
\bibfield{author}{\bibinfo{person}{Xu Liu}, \bibinfo{person}{Juncheng Liu}, \bibinfo{person}{Gerald Woo}, \bibinfo{person}{Taha Aksu}, \bibinfo{person}{Yuxuan Liang}, \bibinfo{person}{Roger Zimmermann}, \bibinfo{person}{Chenghao Liu}, \bibinfo{person}{Silvio Savarese}, \bibinfo{person}{Caiming Xiong}, {and} \bibinfo{person}{Doyen Sahoo}.} \bibinfo{year}{2024}\natexlab{b}.
\newblock \showarticletitle{Moirai-moe: Empowering time series foundation models with sparse mixture of experts}.
\newblock \bibinfo{journal}{\emph{arXiv preprint arXiv:2410.10469}} (\bibinfo{year}{2024}).
\newblock


\bibitem[Loshchilov and Hutter(2017)]%
        {loshchilov2017decoupled}
\bibfield{author}{\bibinfo{person}{Ilya Loshchilov} {and} \bibinfo{person}{Frank Hutter}.} \bibinfo{year}{2017}\natexlab{}.
\newblock \showarticletitle{Decoupled weight decay regularization}.
\newblock \bibinfo{journal}{\emph{arXiv preprint arXiv:1711.05101}} (\bibinfo{year}{2017}).
\newblock


\bibitem[Luo et~al\mbox{.}(2025)]%
        {luo2025time}
\bibfield{author}{\bibinfo{person}{Yucong Luo}, \bibinfo{person}{Yitong Zhou}, \bibinfo{person}{Mingyue Cheng}, \bibinfo{person}{Jiahao Wang}, \bibinfo{person}{Daoyu Wang}, \bibinfo{person}{Tingyue Pan}, {and} \bibinfo{person}{Jintao Zhang}.} \bibinfo{year}{2025}\natexlab{}.
\newblock \showarticletitle{Time Series Forecasting as Reasoning: A Slow-Thinking Approach with Reinforced LLMs}.
\newblock \bibinfo{journal}{\emph{arXiv preprint arXiv:2506.10630}} (\bibinfo{year}{2025}).
\newblock


\bibitem[McCracken and Ng(2016)]%
        {mccracken2016fred}
\bibfield{author}{\bibinfo{person}{Michael~W McCracken} {and} \bibinfo{person}{Serena Ng}.} \bibinfo{year}{2016}\natexlab{}.
\newblock \showarticletitle{FRED-MD: A monthly database for macroeconomic research}.
\newblock \bibinfo{journal}{\emph{Journal of Business \& Economic Statistics}} \bibinfo{volume}{34}, \bibinfo{number}{4} (\bibinfo{year}{2016}), \bibinfo{pages}{574--589}.
\newblock


\bibitem[Nie et~al\mbox{.}(2025)]%
        {nie2025large}
\bibfield{author}{\bibinfo{person}{Shen Nie}, \bibinfo{person}{Fengqi Zhu}, \bibinfo{person}{Zebin You}, \bibinfo{person}{Xiaolu Zhang}, \bibinfo{person}{Jingyang Ou}, \bibinfo{person}{Jun Hu}, \bibinfo{person}{Jun Zhou}, \bibinfo{person}{Yankai Lin}, \bibinfo{person}{Ji-Rong Wen}, {and} \bibinfo{person}{Chongxuan Li}.} \bibinfo{year}{2025}\natexlab{}.
\newblock \showarticletitle{Large language diffusion models}.
\newblock \bibinfo{journal}{\emph{arXiv preprint arXiv:2502.09992}} (\bibinfo{year}{2025}).
\newblock


\bibitem[Nie et~al\mbox{.}(2022)]%
        {nie2022time}
\bibfield{author}{\bibinfo{person}{Yuqi Nie}, \bibinfo{person}{Nam~H Nguyen}, \bibinfo{person}{Phanwadee Sinthong}, {and} \bibinfo{person}{Jayant Kalagnanam}.} \bibinfo{year}{2022}\natexlab{}.
\newblock \showarticletitle{A time series is worth 64 words: Long-term forecasting with transformers}.
\newblock \bibinfo{journal}{\emph{arXiv preprint arXiv:2211.14730}} (\bibinfo{year}{2022}).
\newblock


\bibitem[Oreshkin et~al\mbox{.}(2019)]%
        {oreshkin2019n}
\bibfield{author}{\bibinfo{person}{Boris~N Oreshkin}, \bibinfo{person}{Dmitri Carpov}, \bibinfo{person}{Nicolas Chapados}, {and} \bibinfo{person}{Yoshua Bengio}.} \bibinfo{year}{2019}\natexlab{}.
\newblock \showarticletitle{N-BEATS: Neural basis expansion analysis for interpretable time series forecasting}.
\newblock \bibinfo{journal}{\emph{arXiv preprint arXiv:1905.10437}} (\bibinfo{year}{2019}).
\newblock


\bibitem[Panagopoulos et~al\mbox{.}(2021)]%
        {panagopoulos2021transfer}
\bibfield{author}{\bibinfo{person}{George Panagopoulos}, \bibinfo{person}{Giannis Nikolentzos}, {and} \bibinfo{person}{Michalis Vazirgiannis}.} \bibinfo{year}{2021}\natexlab{}.
\newblock \showarticletitle{Transfer graph neural networks for pandemic forecasting}. In \bibinfo{booktitle}{\emph{Proceedings of the AAAI Conference on Artificial Intelligence}}, Vol.~\bibinfo{volume}{35}. \bibinfo{pages}{4838--4845}.
\newblock


\bibitem[Paszke et~al\mbox{.}(2017)]%
        {paszke2017automatic}
\bibfield{author}{\bibinfo{person}{Adam Paszke}, \bibinfo{person}{Sam Gross}, \bibinfo{person}{Soumith Chintala}, \bibinfo{person}{Gregory Chanan}, \bibinfo{person}{Edward Yang}, \bibinfo{person}{Zachary DeVito}, \bibinfo{person}{Zeming Lin}, \bibinfo{person}{Alban Desmaison}, \bibinfo{person}{Luca Antiga}, {and} \bibinfo{person}{Adam Lerer}.} \bibinfo{year}{2017}\natexlab{}.
\newblock \showarticletitle{Automatic differentiation in pytorch}.
\newblock  (\bibinfo{year}{2017}).
\newblock


\bibitem[{PeMS}({[n.\,d.]})]%
        {trafficdata}
\bibfield{author}{\bibinfo{person}{{PeMS}}.} \bibinfo{year}{[n.\,d.]}\natexlab{}.
\newblock \bibinfo{title}{{Traffic Dataset}}.
\newblock \bibinfo{howpublished}{\url{http://pems.dot.ca.gov/}}.
\newblock


\bibitem[Poyatos et~al\mbox{.}(2016)]%
        {poyatos2016sapfluxnet}
\bibfield{author}{\bibinfo{person}{Rafael Poyatos}, \bibinfo{person}{V{\'\i}ctor Granda}, \bibinfo{person}{Roberto Molowny-Horas}, \bibinfo{person}{Maurizio Mencuccini}, \bibinfo{person}{Kathy Steppe}, {and} \bibinfo{person}{Jordi Mart{\'\i}nez-Vilalta}.} \bibinfo{year}{2016}\natexlab{}.
\newblock \bibinfo{title}{SAPFLUXNET: towards a global database of sap flow measurements}.
\newblock \bibinfo{numpages}{1449--1455}~pages.
\newblock


\bibitem[Qiu et~al\mbox{.}(2024)]%
        {qiu2024tfb}
\bibfield{author}{\bibinfo{person}{Xiangfei Qiu}, \bibinfo{person}{Jilin Hu}, \bibinfo{person}{Lekui Zhou}, \bibinfo{person}{Xingjian Wu}, \bibinfo{person}{Junyang Du}, \bibinfo{person}{Buang Zhang}, \bibinfo{person}{Chenjuan Guo}, \bibinfo{person}{Aoying Zhou}, \bibinfo{person}{Christian~S Jensen}, \bibinfo{person}{Zhenli Sheng}, {et~al\mbox{.}}} \bibinfo{year}{2024}\natexlab{}.
\newblock \showarticletitle{Tfb: Towards comprehensive and fair benchmarking of time series forecasting methods}.
\newblock \bibinfo{journal}{\emph{arXiv preprint arXiv:2403.20150}} (\bibinfo{year}{2024}).
\newblock


\bibitem[Syu and Wang(2021)]%
        {syu2021qos}
\bibfield{author}{\bibinfo{person}{Yang Syu} {and} \bibinfo{person}{Chien-Min Wang}.} \bibinfo{year}{2021}\natexlab{}.
\newblock \showarticletitle{QoS time series modeling and forecasting for Web services: A comprehensive survey}.
\newblock \bibinfo{journal}{\emph{IEEE Transactions on Network and Service Management}} \bibinfo{volume}{18}, \bibinfo{number}{1} (\bibinfo{year}{2021}), \bibinfo{pages}{926--944}.
\newblock


\bibitem[Talukder et~al\mbox{.}(2024)]%
        {talukder2024totem}
\bibfield{author}{\bibinfo{person}{Sabera Talukder}, \bibinfo{person}{Yisong Yue}, {and} \bibinfo{person}{Georgia Gkioxari}.} \bibinfo{year}{2024}\natexlab{}.
\newblock \showarticletitle{Totem: Tokenized time series embeddings for general time series analysis}.
\newblock \bibinfo{journal}{\emph{arXiv preprint arXiv:2402.16412}} (\bibinfo{year}{2024}).
\newblock


\bibitem[Van Den~Oord et~al\mbox{.}(2017)]%
        {van2017neural}
\bibfield{author}{\bibinfo{person}{Aaron Van Den~Oord}, \bibinfo{person}{Oriol Vinyals}, {et~al\mbox{.}}} \bibinfo{year}{2017}\natexlab{}.
\newblock \showarticletitle{Neural discrete representation learning}.
\newblock \bibinfo{journal}{\emph{Advances in neural information processing systems}}  \bibinfo{volume}{30} (\bibinfo{year}{2017}).
\newblock


\bibitem[Wang et~al\mbox{.}(2023)]%
        {wang2023micn}
\bibfield{author}{\bibinfo{person}{Huiqiang Wang}, \bibinfo{person}{Jian Peng}, \bibinfo{person}{Feihu Huang}, \bibinfo{person}{Jince Wang}, \bibinfo{person}{Junhui Chen}, {and} \bibinfo{person}{Yifei Xiao}.} \bibinfo{year}{2023}\natexlab{}.
\newblock \showarticletitle{Micn: Multi-scale local and global context modeling for long-term series forecasting}. In \bibinfo{booktitle}{\emph{The eleventh international conference on learning representations}}.
\newblock


\bibitem[Wang et~al\mbox{.}(2025)]%
        {wang2025can}
\bibfield{author}{\bibinfo{person}{Jiahao Wang}, \bibinfo{person}{Mingyue Cheng}, {and} \bibinfo{person}{Qi Liu}.} \bibinfo{year}{2025}\natexlab{}.
\newblock \showarticletitle{Can slow-thinking llms reason over time? empirical studies in time series forecasting}.
\newblock \bibinfo{journal}{\emph{arXiv preprint arXiv:2505.24511}} (\bibinfo{year}{2025}).
\newblock


\bibitem[Wang et~al\mbox{.}(2024a)]%
        {wang2024deep}
\bibfield{author}{\bibinfo{person}{Yuxuan Wang}, \bibinfo{person}{Haixu Wu}, \bibinfo{person}{Jiaxiang Dong}, \bibinfo{person}{Yong Liu}, \bibinfo{person}{Mingsheng Long}, {and} \bibinfo{person}{Jianmin Wang}.} \bibinfo{year}{2024}\natexlab{a}.
\newblock \showarticletitle{Deep time series models: A comprehensive survey and benchmark}.
\newblock \bibinfo{journal}{\emph{arXiv preprint arXiv:2407.13278}} (\bibinfo{year}{2024}).
\newblock


\bibitem[Wang et~al\mbox{.}(2024b)]%
        {wang2024tssurvey}
\bibfield{author}{\bibinfo{person}{Yuxuan Wang}, \bibinfo{person}{Haixu Wu}, \bibinfo{person}{Jiaxiang Dong}, \bibinfo{person}{Yong Liu}, \bibinfo{person}{Mingsheng Long}, {and} \bibinfo{person}{Jianmin Wang}.} \bibinfo{year}{2024}\natexlab{b}.
\newblock \showarticletitle{Deep Time Series Models: A Comprehensive Survey and Benchmark}.
\newblock  (\bibinfo{year}{2024}).
\newblock


\bibitem[Wang et~al\mbox{.}(2024c)]%
        {wang2024timexer}
\bibfield{author}{\bibinfo{person}{Yuxuan Wang}, \bibinfo{person}{Haixu Wu}, \bibinfo{person}{Jiaxiang Dong}, \bibinfo{person}{Guo Qin}, \bibinfo{person}{Haoran Zhang}, \bibinfo{person}{Yong Liu}, \bibinfo{person}{Yunzhong Qiu}, \bibinfo{person}{Jianmin Wang}, {and} \bibinfo{person}{Mingsheng Long}.} \bibinfo{year}{2024}\natexlab{c}.
\newblock \showarticletitle{Timexer: Empowering transformers for time series forecasting with exogenous variables}.
\newblock \bibinfo{journal}{\emph{arXiv preprint arXiv:2402.19072}} (\bibinfo{year}{2024}).
\newblock


\bibitem[{Wetterstation}({[n.\,d.]})]%
        {weatherdata}
\bibfield{author}{\bibinfo{person}{{Wetterstation}}.} \bibinfo{year}{[n.\,d.]}\natexlab{}.
\newblock \bibinfo{title}{{Weather Dataset}}.
\newblock \bibinfo{howpublished}{\url{https://www.bgc-jena.mpg.de/wetter/}}.
\newblock


\bibitem[Wu et~al\mbox{.}(2022)]%
        {wu2022timesnet}
\bibfield{author}{\bibinfo{person}{Haixu Wu}, \bibinfo{person}{Tengge Hu}, \bibinfo{person}{Yong Liu}, \bibinfo{person}{Hang Zhou}, \bibinfo{person}{Jianmin Wang}, {and} \bibinfo{person}{Mingsheng Long}.} \bibinfo{year}{2022}\natexlab{}.
\newblock \showarticletitle{Timesnet: Temporal 2d-variation modeling for general time series analysis}.
\newblock \bibinfo{journal}{\emph{arXiv preprint arXiv:2210.02186}} (\bibinfo{year}{2022}).
\newblock


\bibitem[Wu et~al\mbox{.}(2021)]%
        {wu2021autoformer}
\bibfield{author}{\bibinfo{person}{Haixu Wu}, \bibinfo{person}{Jiehui Xu}, \bibinfo{person}{Jianmin Wang}, {and} \bibinfo{person}{Mingsheng Long}.} \bibinfo{year}{2021}\natexlab{}.
\newblock \showarticletitle{Autoformer: Decomposition transformers with auto-correlation for long-term series forecasting}.
\newblock \bibinfo{journal}{\emph{Advances in neural information processing systems}}  \bibinfo{volume}{34} (\bibinfo{year}{2021}), \bibinfo{pages}{22419--22430}.
\newblock


\bibitem[Xie et~al\mbox{.}(2024)]%
        {xie2024show}
\bibfield{author}{\bibinfo{person}{Jinheng Xie}, \bibinfo{person}{Weijia Mao}, \bibinfo{person}{Zechen Bai}, \bibinfo{person}{David~Junhao Zhang}, \bibinfo{person}{Weihao Wang}, \bibinfo{person}{Kevin~Qinghong Lin}, \bibinfo{person}{Yuchao Gu}, \bibinfo{person}{Zhijie Chen}, \bibinfo{person}{Zhenheng Yang}, {and} \bibinfo{person}{Mike~Zheng Shou}.} \bibinfo{year}{2024}\natexlab{}.
\newblock \showarticletitle{Show-o: One single transformer to unify multimodal understanding and generation}.
\newblock \bibinfo{journal}{\emph{arXiv preprint arXiv:2408.12528}} (\bibinfo{year}{2024}).
\newblock


\bibitem[Xue and Salim(2023)]%
        {xue2023promptcast}
\bibfield{author}{\bibinfo{person}{Hao Xue} {and} \bibinfo{person}{Flora~D Salim}.} \bibinfo{year}{2023}\natexlab{}.
\newblock \showarticletitle{Promptcast: A new prompt-based learning paradigm for time series forecasting}.
\newblock \bibinfo{journal}{\emph{IEEE Transactions on Knowledge and Data Engineering}} \bibinfo{volume}{36}, \bibinfo{number}{11} (\bibinfo{year}{2023}), \bibinfo{pages}{6851--6864}.
\newblock


\bibitem[Zeng et~al\mbox{.}(2023)]%
        {zeng2023transformers}
\bibfield{author}{\bibinfo{person}{Ailing Zeng}, \bibinfo{person}{Muxi Chen}, \bibinfo{person}{Lei Zhang}, {and} \bibinfo{person}{Qiang Xu}.} \bibinfo{year}{2023}\natexlab{}.
\newblock \showarticletitle{Are transformers effective for time series forecasting?}. In \bibinfo{booktitle}{\emph{Proceedings of the AAAI conference on artificial intelligence}}, Vol.~\bibinfo{volume}{37}. \bibinfo{pages}{11121--11128}.
\newblock


\bibitem[Zhang et~al\mbox{.}(2021)]%
        {zhang2021time}
\bibfield{author}{\bibinfo{person}{Lingyu Zhang}, \bibinfo{person}{Wenjie Bian}, \bibinfo{person}{Wenyi Qu}, \bibinfo{person}{Liheng Tuo}, {and} \bibinfo{person}{Yunhai Wang}.} \bibinfo{year}{2021}\natexlab{}.
\newblock \showarticletitle{Time series forecast of sales volume based on XGBoost}. In \bibinfo{booktitle}{\emph{Journal of Physics: Conference Series}}, Vol.~\bibinfo{volume}{1873}. IOP Publishing, \bibinfo{pages}{012067}.
\newblock


\bibitem[Zhang et~al\mbox{.}(2023)]%
        {zhang2023copilot4d}
\bibfield{author}{\bibinfo{person}{Lunjun Zhang}, \bibinfo{person}{Yuwen Xiong}, \bibinfo{person}{Ze Yang}, \bibinfo{person}{Sergio Casas}, \bibinfo{person}{Rui Hu}, {and} \bibinfo{person}{Raquel Urtasun}.} \bibinfo{year}{2023}\natexlab{}.
\newblock \showarticletitle{Copilot4d: Learning unsupervised world models for autonomous driving via discrete diffusion}.
\newblock \bibinfo{journal}{\emph{arXiv preprint arXiv:2311.01017}} (\bibinfo{year}{2023}).
\newblock


\bibitem[Zhou et~al\mbox{.}(2021)]%
        {zhou2021informer}
\bibfield{author}{\bibinfo{person}{Haoyi Zhou}, \bibinfo{person}{Shanghang Zhang}, \bibinfo{person}{Jieqi Peng}, \bibinfo{person}{Shuai Zhang}, \bibinfo{person}{Jianxin Li}, \bibinfo{person}{Hui Xiong}, {and} \bibinfo{person}{Wancai Zhang}.} \bibinfo{year}{2021}\natexlab{}.
\newblock \showarticletitle{Informer: Beyond efficient transformer for long sequence time-series forecasting}. In \bibinfo{booktitle}{\emph{Proceedings of the AAAI conference on artificial intelligence}}, Vol.~\bibinfo{volume}{35}. \bibinfo{pages}{11106--11115}.
\newblock


\bibitem[Zhou et~al\mbox{.}(2022a)]%
        {zhou2022film}
\bibfield{author}{\bibinfo{person}{Tian Zhou}, \bibinfo{person}{Ziqing Ma}, \bibinfo{person}{Qingsong Wen}, \bibinfo{person}{Liang Sun}, \bibinfo{person}{Tao Yao}, \bibinfo{person}{Wotao Yin}, \bibinfo{person}{Rong Jin}, {et~al\mbox{.}}} \bibinfo{year}{2022}\natexlab{a}.
\newblock \showarticletitle{Film: Frequency improved legendre memory model for long-term time series forecasting}.
\newblock \bibinfo{journal}{\emph{Advances in neural information processing systems}}  \bibinfo{volume}{35} (\bibinfo{year}{2022}), \bibinfo{pages}{12677--12690}.
\newblock


\bibitem[Zhou et~al\mbox{.}(2022b)]%
        {zhou2022fedformer}
\bibfield{author}{\bibinfo{person}{Tian Zhou}, \bibinfo{person}{Ziqing Ma}, \bibinfo{person}{Qingsong Wen}, \bibinfo{person}{Xue Wang}, \bibinfo{person}{Liang Sun}, {and} \bibinfo{person}{Rong Jin}.} \bibinfo{year}{2022}\natexlab{b}.
\newblock \showarticletitle{Fedformer: Frequency enhanced decomposed transformer for long-term series forecasting}. In \bibinfo{booktitle}{\emph{International conference on machine learning}}. PMLR, \bibinfo{pages}{27268--27286}.
\newblock


\bibitem[Zhu et~al\mbox{.}(2024)]%
        {zhu2024addressing}
\bibfield{author}{\bibinfo{person}{Yongxin Zhu}, \bibinfo{person}{Bocheng Li}, \bibinfo{person}{Yifei Xin}, {and} \bibinfo{person}{Linli Xu}.} \bibinfo{year}{2024}\natexlab{}.
\newblock \showarticletitle{Addressing representation collapse in vector quantized models with one linear layer}.
\newblock \bibinfo{journal}{\emph{arXiv preprint arXiv:2411.02038}} (\bibinfo{year}{2024}).
\newblock


\end{thebibliography}
